\documentclass[pmlr]{jmlr} 

\newcommand{\tagdsmode}{proceedings}

\makeatletter
\newcommand{\tagdssubmission}{submission}
\newcommand{\tagdsproceedings}{proceedings}

\ifx\tagdsmode\tagdsproceedings

\else\ifx\tagdsmode\tagdssubmission
  \def\ps@jmlrtps{%
    \let\@mkboth\@gobbletwo
    \def\@oddhead{\scriptsize Under Review at the 2nd Conference on Topology, Algebra, and Geometry in Data Science\hfill}%
    \let\@evenhead\@oddhead
    \def\@oddfoot{}%
    \let\@evenfoot\@oddfoot
  }

\else
  \def\ps@jmlrtps{%
    \let\@mkboth\@gobbletwo
    \def\@oddhead{}%
    \let\@evenhead\@oddhead
    \def\@oddfoot{}%
    \let\@evenfoot\@oddfoot
  }
\fi\fi
\makeatother

\usepackage{longtable}

\usepackage{booktabs}
\usepackage[load-configurations=version-1]{siunitx} 

\theorembodyfont{\upshape}
\theoremheaderfont{\scshape}
\theorempostheader{:}
\theoremsep{\newline}

\jmlrvolume{334}
\jmlryear{2026}
\jmlrworkshop{Topology, Algebra, and Geometry in Data Science}
\jmlrpages{364--386}

\title[Symmetry Learning via Gradient Alignment ]{
Loss Landscape Geometry of Partial Differential Equation Emulators: Or, Symmetry Learning via Gradient Alignment 
}

\ifx\tagdsmode\tagdssubmission

\else

\author{%
  \Name{James Amarel} \Email{jlamarel@lanl.gov} \\
  \Name{Robyn Miller} \Email{robynm@lanl.gov}\\
  \Name{Nicolas Hengartner} \Email{nickh@lanl.gov}\\
  \Name{Benjamin Migliori} \Email{ben.migliori@lanl.gov}\\
  \Name{Emily Taylor} \Email{ecasleton@lanl.gov}\\
  \Name{Alexei Skurikhin} \Email{alexei@lanl.gov}\\
  \Name{Earl Lawrence} \Email{earl@lanl.gov}\\
  \Name{Gerd J. Kunde} \Email{g.j.kunde@lanl.gov}\\
  \addr Los Alamos National Laboratory, Los Alamos, NM 87545
}

\fi

\begin{document}

\maketitle

\begin{abstract}
We study how neural emulators of partial differential equation solution operators learn
physical symmetries from data by introducing a hat-matrix diagnostic that quantifies the
alignment of parameter updates between symmetry related training examples. 
The diagnostic is a metric-weighted overlap of loss gradients evaluated across group orbits, giving a proximal influence function for symmetry-related examples. 
Our measurements of gradient alignment  across both
translations and rotations for models trained as autoregressive fluid-flow emulators suggest
that equivariance arises when training updates propagate gradients coherently throughout
symmetry orbits. This finding is based on an empirical correspondence between equivariance
error and cross-orbit influence. Both our UNet and ViT architectures exhibit approximate
translation equivariance, yet their gradient alignment profiles differ by uniform versus
periodically concentrated influence over the orbit. On a Navier-Stokes dataset, pronounced
dihedral equivariance error coincides with suppressed cross-influence, identifying the
failure as due to decoupled learning of symmetry group elements.  Our diagnostic is
architecture-agnostic and probes a mechanism for learning symmetries from data, extending
beyond forward-pass equivariance tests by directly assessing whether learning updates share
information across physically equivalent configurations. We identify symmetry generalization
performance with symmetry-compatible gradient transport, suggesting that evaluation of
scientific machine learning models requires dynamical probes of loss landscape geometry in
addition to predictive accuracy.
\end{abstract}

\section{Introduction} 
Deep learning emulators for partial differential equation (PDE) solvers routinely achieve
impressive in-distribution accuracy \citep{brandstetter2023messagepassingneuralpde,
herde2024poseidonefficientfoundationmodels,
takamoto2024pdebenchextensivebenchmarkscientific, lippe2023pderefinerachievingaccuratelong,
gupta2022multispatiotemporalscalegeneralizedpdemodeling,
ohana2025welllargescalecollectiondiverse}, yet they often fail to respect the fundamental
symmetries of the governing equations \citep{akhoundsadegh2023liepointsymmetryphysics,
gregory2024equivariant, gruver2022}. This limitation undermines
their ability to extrapolate and generalize, raising the question: are such models truly
learning physics, or merely fitting correlations present in the training data?  Explaining
this gap requires probing not just the outputs, but also the learning process
\citep{fort2020stiffnessnewperspectivegeneralization, zhao2024improvingconvergence}.

Symmetries of the Navier-Stokes (NS) equations, namely translations, rotations, reflections,
scalings, and Galilean boosts, organize the solution space into orbits whose members are
physically equivalent \citep{pmlr-v162-brandstetter22a}. A model that has learned the
solution operator will share information across these orbits: gradients of the loss with
respect to parameters, evaluated on symmetry related inputs, should align, on account of
equivariance; without such coherence, the resulting loss differentials do not constructively
influence one another, rendering the orbit decoupled. When gradients constructively couple across examples,
the respective losses are reduced sympathetically. Measuring cross-influence thus offers a
diagnostic beyond standard forward-pass equivariance checks, exposing the degree to which
gradient signals reflect physical expectations.

If influence across group actions presents only weakly, the model is memorizing localized
patterns \citep{arpit2017closerlookmemorizationdeep, he2020localelasticityneuralnetworks,
chatterjee2020coherentgradientsapproachunderstanding} rather than learning physical
processes. 
Conversely, marked gradient alignment demonstrates that the network has learned to couple
symmetry related states, consistent with the behavior of a true solution operator. Our
symmetry-aware gradient diagnostic therefore probes a process by which models learn to generalize
across orbits, providing a principled tool to assess how architectural choices, loss design,
and inductive biases promote, or hinder, generalization.

We study models that have learned approximate translation equivariance, despite the data
distribution of initial conditions not being stationary under translations. This data-driven
attainment of generalization across the translation group is considered together with 
a similar setup involving the square group; the NS dataset that we consider
presents initial conditions that are invariant only under the Klein four-subgroup of the
square group, and it is only these symmetry transformations that are learned by our
models. By considering the structural correspondence of influence profiles and equivariance
error, our work contributes to better understanding the underlying process by which
physical symmetries can be learned from data distributions that are not manifestly invariant
to the associated symmetry group. It is important to recognize that a test example can be both out of distribution
in the statistical sense and physically equivalent to an example represented by said data distribution;
we use the influence to measure if gradient update signals leverage this physical equivalence.

\section{Related Work}

Encoding symmetries as inductive biases has a long tradition in geometric deep learning.
Group-equivariant CNNs generalize convolution to arbitrary groups with weight sharing across
orbits \citep{CohenWelling2016}. Steerable CNNs make these ideas explicit by parameterizing
kernels in group-steerable bases, including variants for volumetric data
\citep{CohenWelling2017,Weiler2018}. For non-grid domains, tensor field networks
\citep{Thomas2018} and E(n)-equivariant graph neural networks \citep{Satorras2021} extend
manifest symmetry compliance to point clouds and molecules. In practice, scientific data
rarely obey exact symmetries; boundary conditions, material inhomogeneities, grid
discretization, and measurement noise introduce systematic symmetry breaking. This motivates
research into approximate \citep{wang2022approximatelyequivariantnetworksimperfectly} and
relaxed \citep{finzi2021} attainment of equivariance. Furthermore, many contemporary
large-scale models \citep{climax, aurora} rely on data-driven learning of symmetry.
Empirical and theoretical results substantiate this strategy: softening hard constraints can
improve optimization and accuracy 
\citep{wang2022approximatelyequivariantnetworksimperfectly}. Moreover, classic studies show
that modern CNNs can be brittle to small shifts and rotations 
\citep{JMLR:v20:19-519,pmlr-v97-zhang19a,kayhan2020translationinvariancecnnsconvolutional}.

While a constructive route for guaranteed equivariance exists, this is not the regime
targeted by modern large-scale scientific models. Indeed, contemporary state-of-the-art
models for PDE emulation and weather forecasting largely use architectures comprised of
transformer and convolutional layers that are not equivariant by design
\citep{takamoto2024pdebenchextensivebenchmarkscientific,
    lippe2023pderefinerachievingaccuratelong,
    gupta2022multispatiotemporalscalegeneralizedpdemodeling,
    ohana2025welllargescalecollectiondiverse, herde2024poseidonefficientfoundationmodels,
walrus, aurora, climax}. Our focus is therefore on settings in which
symmetry learning is data-driven and contingent on local update geometry rather than
guaranteed a-priori.  The restriction to exactly equivariant models is not necessarily
optimal when realistic data and optimization difficulties are considered \citep{wang2022approximatelyequivariantnetworksimperfectly}. Moreover, for
architectures that enforce exact equivariance by construction, influence across symmetry
orbits would be uniform by design. The purpose of this work is expressly to better
understand how models learn symmetry when the inductive bias is not explicitly given to the
model through, e.g., architecture design, training losses, or data augmentation.

In cases where a model lacking manifest symmetry is selected, probes are needed to
both quantify and explain the degree of equivariance achieved after training. Several
diagnostics test the extent to which a model has learned symmetry, e.g., forward‑pass checks to evaluate
equivariance error under group actions \citep{equivdyna_2024, xie2025talesymmetriesexploringloss},
characterizations of internal representations \citep{tegan},
and the Lie derivative metric, which quantifies
infinitesimal equivariance with layerwise decomposition \citep{gruver2022}. 
Yet,
spot tests with these metrics are insufficient to explain
the mechanism underlying symmetry learning.
Such diagnostics characterize equivariance of the learned map, but do not 
probe processes underlying data-driven symmetry-learning.
In contrast, our work relates observed equivariance behavior to architectural choices and training dynamics
by measuring the propagation of information across symmetry related states.

A
central approach in explainable artificial intelligence is to interrogate the loss and its
derivatives to connect predictive behavior with training signals.
Related work on gradient geometry links cross‑example
parameter update structure to out‑of‑sample performance: stiffness and coherent
gradients capture alignment, and local elasticity studies stability under SGD
updates on distant samples 
\citep{fort2020stiffnessnewperspectivegeneralization,chatterjee2020coherentgradientsapproachunderstanding,he2020localelasticityneuralnetworks}. 
Euclidean influence functions trace predictions to training data but are delicate in deep, non‑convex regimes, motivating curvature‑aware variants 
\citep{koh2020understandingblackboxpredictionsinfluence,BasuEtAl2020, jacot2020neuraltangentkernelconvergence,fort2019emergentpropertieslocalgeometry}.

\section{Method}
We compare a UNet (13M parameters, 4 down-sampling blocks, 24 embedding channels) and a
Vision Transformer (ViT; 5M parameters, 6 layers, 256 channels) trained as emulators for
two-dimensional compressible Euler flows from PDEGym
\citep{herde2024poseidonefficientfoundationmodels}. For training data, we selected \(3\) classes of
Riemann-type initial conditions (CE-RP, CE-RPUI, CE-CRP), each with $6{,}500$ trajectories
of 16 time steps. Each state snapshot is a $128\times128$ grid of mass density, Cartesian
momentum density, and energy density. Models were trained autoregressively to emulate the
Euler evolution operator; for more details see our companion paper that considers time-translation symmetry, coherence over initial condition classes, and influence across physics-informed objectives \citep{paper1}.
To complement the compressible Euler results, we also apply our analysis to models trained on velocity fields
generated by the Navier-Stokes (NS) equations using, NS-BB, NS-Gauss, and NS-Sines initial
conditions. These flows occupy a qualitatively different feature space, characterized by
smoother, viscosity-regularized dynamics and vorticity-dominated structure.

Optimization trajectories for CE and NS are shown in \autoref{fig:opt_curves};
training is halted in the late-stage learning regime, so the hat-matrix
diagnostics probe local gradient geometry after the models have entered
low-test-loss basins but before learning has completely saturated. We used Adam
with learning rate $5\times10^{-4}$ and weight decay $\lambda = 10^{-4}$ on
mini-batches of $N=48$ transitions. The cost function was a scaled mean-squared
error (SMSE), that normalizes errors by channel RMS to balance large and
small-amplitude features, supporting the capture of shocks and wavefronts while
retaining sensitivity to quiescent flows, in addition to rendering
dimensionless the influence matrix of interest. Both models were trained in
distributed mode on two 40GB A100 GPUs using Lux.jl \citep{pal2023lux,
pal2023efficient} and Zygote.jl \citep{Zygote.jl-2018}, with \(3\) seeds
controlling initialization and dataset splits. Results are reported with
quantile range bars to capture variability across seeds.

To evaluate our models, we compute the influence function, which can be expressed as the Lie
derivative of the cost along gradient flow induced by individual test examples.  By
formulating the influence through the Lie derivative, we elevate the diagnostic from its
traditional role as a static, final-state, measurement, to a dynamical quantity that can be
meaningfully observed during training. 
However, due to computational expense, the present paper primarily measures
influence at a single parameter point in the late-stage learning regime [see
\autoref{fig:opt_curves}]. The checkpointed evolution of influence is shown in
\autoref{fig:Klein_NS}, \autoref{fig:D4_evolution_NS}, and
\autoref{fig:D4_evolution_CE}.
Let
\(
V^{\mu} = -\chi^{\mu\nu}\partial_{\nu} C_x
\)
denote the vector field generated by the loss evaluated on an example \(x\). The influence
of this update on the loss evaluated at the group-transformed input \(gx\) is given
by\footnote{Programmatically, the translation group action is realized via cyclic
    permutation of array indices. For scalar fields, the square group can be
    effected
    with standard commands for rotating and flipping images; for the velocity field one must
take care to appropriately mix vector indices in addition to rotating the coordinate basis.}
\begin{equation}
    -\mathcal{L}_V C_{gx}
    =
    (\partial_\mu C_{gx})\,\chi^{\mu\nu}\,(\partial_\nu C_{x}),
    \label{eq:influence}
\end{equation}
where the regularized Gauss-Newton metric \(\chi_{\mu \nu} = \eta_{\mu \nu} + \lambda
\delta_{\mu \nu}\)  is comprised of both an isotropic regularization term of strength
\(\lambda\) and \(\eta\), the pullback of the Euclidean metric on function space to
parameter space. Einstein summation convention is implicit and we use standard index raising
notation from differential geometry; \(\chi^{\mu \nu}\) denote the elements of \(\chi^{-1}\)
\citep{absil2008}. \autoref{eq:influence} recovers the familiar influence function
definition as a metric-weighted overlap between gradients derived from the cost evaluated on
an example \(x\) and the transformed counterpart \(g x\)
\citep{fort2019emergentpropertieslocalgeometry}. Intuitively, the influence function
measures whether a gradient update induced by one example decreases (or increases) the loss
of another. When applied to symmetry related inputs, it quantifies whether learning signals
propagate coherently along symmetry orbits. In regression, the Gauss-Newton matrix plays the
role of the Fisher-information by supplying the Jacobian-induced metric on parameter space
\citep{JMLR:v21:17-678}; sandwiching the metric between Jacobians yields the
(\(\lambda\)-regularized) hat-matrix \(J_{\mu}\chi^{\mu \nu}J_{\nu}\), which reduces to the
neural tangent kernel in the Euclidean limit \citep{huber2009robust,
jacot2020neuraltangentkernelconvergence}. The influence function (\autoref{eq:influence}) can
also be expressed by sandwiching the hat-matrix between prediction residuals \citep{paper1}.

For each
model seed, the influence function is evaluated across \(18\) test mini-batches comprising
\(3\) distinct initial conditions over \(16\) time-steps.  
As a first-order sensitivity of a scalar loss to an example-induced update, the influence function has no canonical scale.
Common choices include normalization by gradient magnitude, yielding the cosine similarity,
normalization by trace or Frobenius norms, standardized influence functions
\citep{heritier1994robust, lu1997standardized}, or alternatively leaving the response
unnormalized altogether. In the experiments below, we use cosine normalization, reporting
influence after division by both the perturbation and response gradient magnitudes. This
normalization sets each self-response to unity, which removes the absolute scale of local
sensitivities and hence isolates the cross-response magnitude, a comparison appropriate when
the effects of influence are accumulated over many optimization steps.

For further discussion of data selection, model selection, an intuitive derivation of the influence function,
and details on its computation,
see \autoref{app:data}, \autoref{app:model}, \autoref{app:inf}, and \autoref{app:qr}, respectively.

\section{Results}
Our evaluation jointly considers equivariance error, which probes forward-pass consistency
under symmetry transformations, and influence function matrix elements. The influence
function reveals whether learning  propagates information coherently across
symmetry related states, exposing whether a model is genuinely learning symmetries or merely
fitting correlations in the data.  Convergence to basins that respect the symmetries of the
underlying problem is both essential for generalization and a persistent challenge for
current architectures. Throughout this section, we test the central hypothesis that
forward-pass equivariance emerges when local update geometry propagates gradient information
coherently along symmetry orbits.  We find that forward-pass equivariance error is in structural correspondence with
the extent to which our models and training regimes support symmetry-coherent gradients.

\subsection{Translation Group}
For the translation group, we first evaluate purely horizontal and vertical translations across
the periodic spatial domain. Because the governing PDE applies uniformly in space, the
learned dynamics should be equivariant under translations. In any given flow snapshot,
nonlinear interactions occupy only a small number of localized regions, yet these structures
may arise at arbitrary spatial locations. A model that learns translational equivariance
will therefore treat such interactions consistently wherever they occur, a prerequisite for
both capturing rare but dynamically significant events and enabling accurate long-time
extrapolation.

Both architectures exhibit median relative equivariance error on the order of one percent or
less, indicating that training largely confers translational symmetry generalizing
capabilities. To capture the structure of those examples for which low equivariance error was
not obtained, we consider the third quartile error-profile.
Notably, the relative equivariance error attained by our models [see
\autoref{fig:ZH_CE}] reflects genuine generalization over the translation group: we did
not explicitly augment the training data with translated states, nor was either architecture
designed to enforce exact equivariance over the entire translation group under
consideration. In particular, translation stationarity is violated by the distribution from
which four quadrant initial conditions for both CE-RP and CE-RPUI flows were drawn.
Nevertheless, we observe nontrivial influence across translated states, indicating
that training updates couple translation-related inputs, even without hard
symmetry constraints. 

Inspection of forward-pass equivariance error together with influence profile [see \autoref{fig:ZH_CE}]
provides an explanatory correspondence for symmetry learning: resonant minima in the error
profile coincide with structured extrema in the response, and vice-versa. Both models
support cross-influence reflective of response geometry that supports nontrivial propagation
across the translation orbit and thus generalization over translations is achieved, even in
the absence of explicit translation augmentation or exact equivariance constraints.

\begin{figure}[htbp]
\floatconts
  {fig:ZH_CE}
  {\caption{
Diagnostics under horizontal translations on CE data.
          Markers and rangebars indicate the median and interquartile range over seeds.
          Lack of uniformity reflects horizontal symmetry breaking.
          For vertical translations, see \autoref{fig:ZV_CE}; for analogous results with NS data, see \autoref{fig:ZH_NS}.
}}
  {%
    \setlength{\jmlrminsubcaptionwidth}{0.40\linewidth}%
    \subfigure[
Relative equivariance error as a function of horizontal translation.
    ][t]{%
      \includegraphics[width=0.40\linewidth]{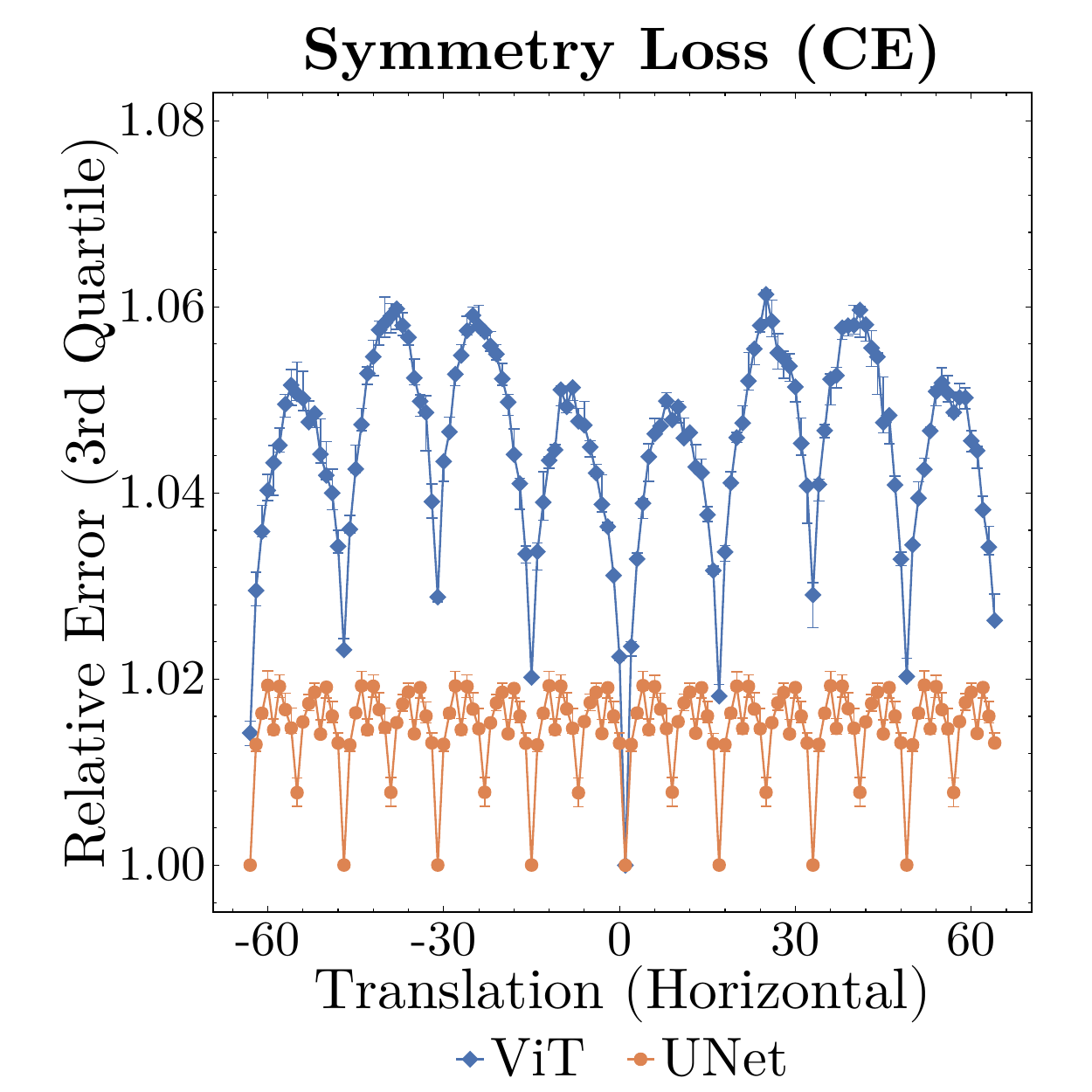}}%
    \hfill
    \subfigure[
 Gradient alignment between an example and its horizontally translated state.
    ][t]{%
      \includegraphics[width=0.40\linewidth]{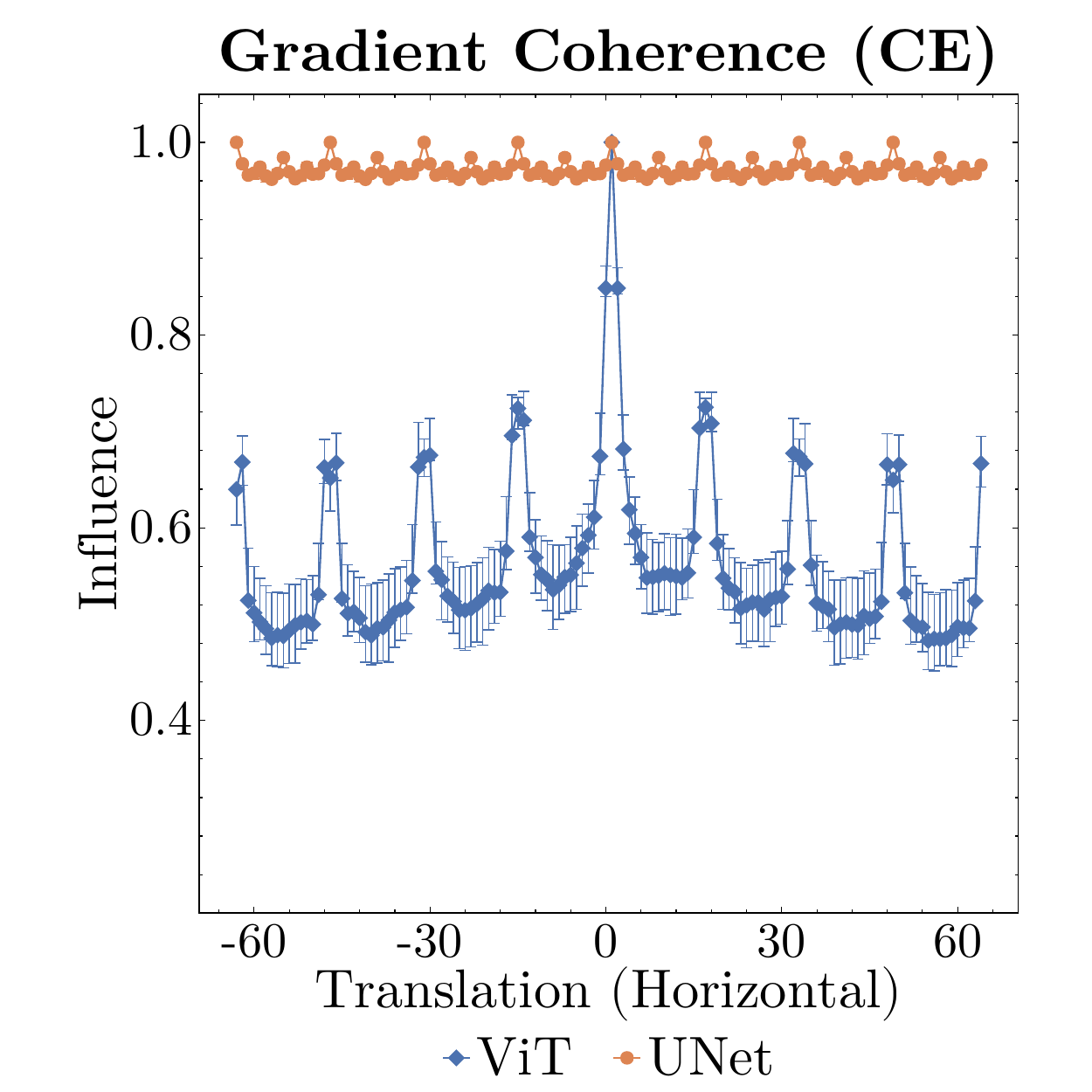}}
  }%
\end{figure}

The translation response for our ViTs [see \autoref{fig:ZH_CE} and
\autoref{fig:ZV_CE}] exhibits a pronounced axis-dependent resonance structure. Under
horizontal (vertical) translations, the periodicity of wavelength $16$ ($64$) pixels
indicates that the measured influence/equivariance observables couple anisotropically to the
phase of the translation relative to the patch lattice, with different group elements
emphasized in different directions. This axis-dependent anisotropy may be due to
representational bias introduced by flattening the two-dimensional patch grid into a
one-dimensional token sequence. While the patch embedding itself is convolutional of stride
four, the subsequent permutation into tokens and channel mixing allows for a preferred
ordering on spatial degrees of freedom, thereby breaking the manifest equivalence between
horizontal and vertical directions, which impacts the learned response geometry. In
contrast, our UNets employ convolutional operators across contracting and expanding
resolutions, with four down-sampling stages. The coarsest representation has spatial extent
$8 \times 8$, which promotes a comparatively smooth dependence of influence on translation,
with only low-amplitude variation at shorter $8$ pixel wavelengths and their subharmonics.
Consistent with this architecture-level partial symmetry adherence, UNets exhibit similar
influence profiles across horizontal and vertical translations.

The joint horizontal and vertical translation grids in
\autoref{fig:box_CE_ViT}, \autoref{fig:box_CE_UNet},
\autoref{fig:box_NS_ViT}, and \autoref{fig:box_NS_UNet} resolve the normalized
first-order loss reduction induced across the sampled two-dimensional
translation orbit by an update from the untransformed example. On CE data, the
ViT retains horizontal and vertical phase bands, whereas the UNet exhibits an
approximately eight-pixel lattice; in both cases, elevated third-quartile
relative equivariance error coincides with reduced gradient coherence. The UNet
exhibits the same complementary organization on NS data, while the ViT--NS grid
gives the weakest correspondence: its error is phase-banded, but its coherence
decays approximately isotropically away from the identity translation.

That our UNets exhibit nearly uniform gradient coherence, while our
ViTs distributes influence non-uniformly yet support larger influence on privileged group
elements, suggests a trade-off between inductive bias and optimization flexibility.
Non-uniform coupling over the group allows our ViTs to concentrate each learning signal on a
subset of translation phases, producing stronger local responses at the cost of more
heterogeneous orbit-wise coupling. Less rigid coupling between translation elements for the
ViT may be advantageous when paired with an overall larger peak response, as its updates are
flexibly dispersed throughout the group. 
Indeed, low cross-influence in the Gauss--Newton geometry indicates that an
update induced by one example produces little first-order change in the loss
of another example.
Enforcing uniform coupling across all group elements may constrain the space of admissible
update directions and, in some settings, impede optimization \citep{pmlr-v97-zhang19a,
JMLR:v20:19-519, kayhan2020translationinvariancecnnsconvolutional} by inducing a mean
gradient direction that is inconsistent with individual update directions \citep{gradpinn}.
In our translation group related experiments, we observe exclusively positive median
influence values, suggesting that optimization is not frustrated by conflicting updates;
instead, the distinction lies in how learning signal is distributed: architectures with
weaker inductive bias are free to concentrate influence on a subset of symmetry
transformations, enabling rapid training loss reduction via a strong gradient response, but
limiting generalization capabilities across the group. 

\subsection{Dihedral Group}
Next, we analyze symmetry learning with respect to the dihedral group $D_4$, the set of
discrete rotations and reflections admissible under the datasets used for training. Our
models generalize on those group transformations for which gradients are
propagated coherently through the loss landscape. Where generalization fails, we show that
the local update geometry does not support full orbit-wise coupling and visualize how data-induced
anisotropies are absorbed into the learned loss landscape geometry.

Our Navier-Stokes trained models fail to learn the dihedral group $D_4$, a breakdown that
presents across random seeds and mini-batches despite otherwise excellent pointwise accuracy
and low untransformed test loss [see \autoref{fig:D4_NS}]. Both architectures achieve
competitive performance on the identity element \(e\), \(180\) degree rotations \(r^2\),
\(90\) degree rotations followed by a reflection about the vertical axis \(sr\), and \(270\)
degree rotations followed by such a reflection \(sr^3\), yet their response under the remaining
group actions reveals a sharp and systematic lack of symmetry adherence. In this case, the
relative SMSE increases catastrophically, by nearly $10^{4}$, indicating that the learned
predictors are not even approximately equivariant when all group actions are considered.
Such behavior is not entirely unexpected. 
The Navier--Stokes dataset carries a pronounced directional bias in feature
space, inherited from the choice of initial conditions and the anisotropic,
vorticity-dominated structure of the flows: the data distribution is invariant,
and our models are approximately equivariant, under the Klein four-subgroup
\(V_4=\{e,r^2,sr,sr^3\}\), but not under the complementary actions
\(r,r^3,s,sr^2\). Such transformed states are physically admissible under
reasonable deployment assumptions, so the observed decoupling constitutes a
generalization gap between the training and operational distributions.
\footnote{We thank Reviewer q6u2 for naming the Klein four-subgroup and sharpening our
    discussion on the role of the initial condition invariance group. Our results show that
    the invariance group is not the sole factor in determining generalization capabilities.
    Indeed, while our models fail to generalize across \(D_4\), they attain approximate
    equivariance over translations. In both cases, the initial-condition distribution is not invariant under the
respective symmetry group; a distinguishing observable between these two
scenarios is the cross-influence profile.
}
What is striking, however, is that the symmetry
breaking is not merely a property of the forward map. 
The influence structure associated with these unlearned group actions is
distinguished in \autoref{fig:D4_NS}: the group elements with catastrophic
equivariance error are precisely those with suppressed cross-influence.
\begin{figure}[htbp]
\floatconts
  {fig:D4_NS}
  {\caption{
Diagnostics under \(D_{4}\) symmetry on Navier-Stokes (NS) data.
          Markers and rangebars indicate the median and interquartile range over seeds.
          Nonuniformity across group elements reflects dihedral symmetry breaking; group
          actions with large equivariance error exhibit suppressed gradient alignment.
		For the corresponding CE diagnostics, see \autoref{fig:D4_CE}.
  }}
  {%
    \setlength{\jmlrminsubcaptionwidth}{0.40\linewidth}%
    \subfigure[
 Relative equivariance error for each \(D_{4}\) group action.
    ][t]{%
      \includegraphics[width=0.40\linewidth]{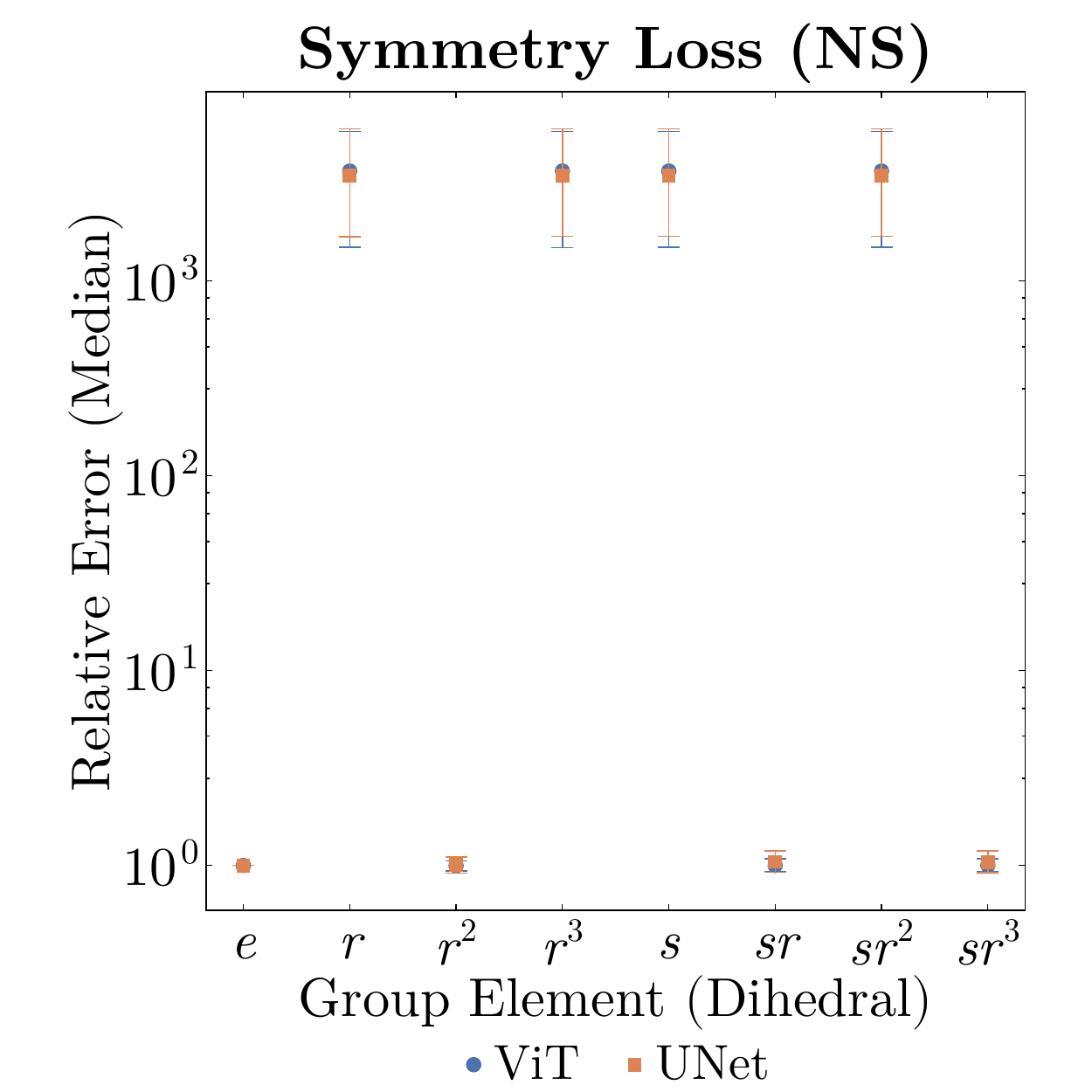}}%
    \hfill
    \subfigure[
Gradient alignment between an example and its \(D_{4}\)-transformed state.
][t]{%
      \includegraphics[width=0.40\linewidth]{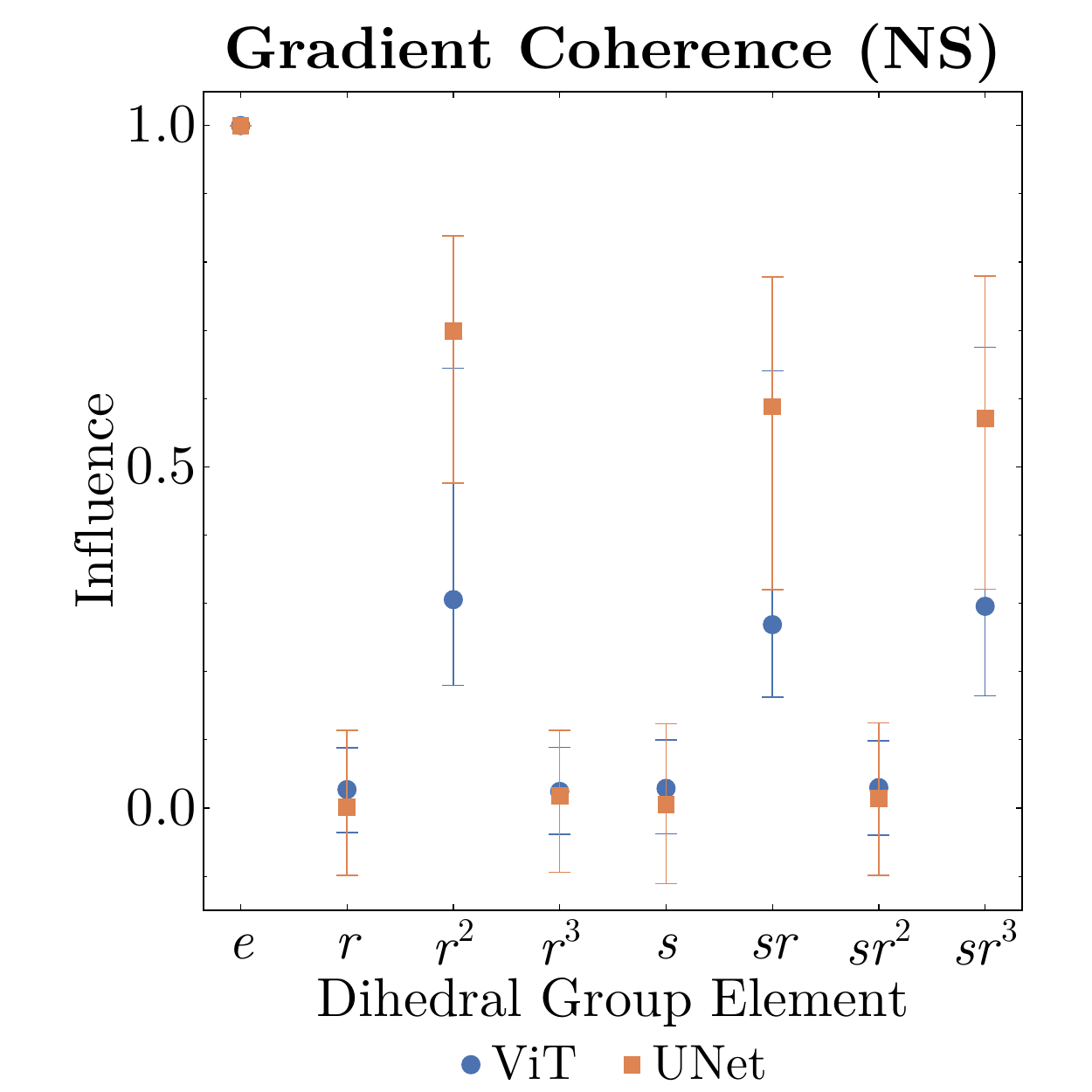}}
  }%
\end{figure}

Rather than facilitating a democratic sharing of coupling across the dihedral orbit,
optimization drove our models into a region of the loss landscape that displays a clear
demarcation between well-learned and challenging group elements, with the latter receiving
subdominant influence. These challenging group elements receive near-zero
cross-influence, indicating that training has converged to a basin
with symmetry-incompatible local geometry. 
The checkpointed influence profiles in \autoref{fig:Klein_NS} reveal how this
orbit-wise separation develops throughout optimization. Although the two
sectors begin with comparable cross-influence, both UNet and ViT rapidly
establish and maintain stronger transfer within \(V_4\setminus e\) than into
\(D_4\setminus V_4\), yielding training dynamics that support coherent gradient
propagation over the learned Klein four-subgroup but not the full dihedral
orbit.

We emphasize that although the dihedral symmetry
is physically valid, gradient
signals do not accumulate coherently over symmetry operations. Instead, data-induced
anisotropies are absorbed into the local loss geometry, inducing local updates that
reinforce symmetry-breaking. Our influence probe
interrogates directions that are poorly aligned with the primary drift direction traced by
the training flow. Hence, symmetry related gradient signals enter only as subleading
perturbations and do not accumulate coherently, explaining how models achieve excellent
pointwise accuracy on in-training-distribution data while converging to basins whose local
geometry explicitly breaks dihedral symmetry.
Indeed, \autoref{fig:D4_NS} explains the apparent tension between low
in-training-distribution test error and severe dihedral failure: under
symmetry-agnostic training, the induced local update geometry does not
propagate information coherently across the full \(D_4\) orbit, so neither
model class learns the \(D_4\)-equivariant extension expected of the underlying
solution operator. 

\section{Conclusion}
Our analysis sharpens recent insights on the interplay between inductive bias,
optimization, and symmetry learning \citep{equivdyna_2024}. We show that test
time equivariance error  tracks how the local update geometry distributes influence
across group orbits. While symmetry-agnostic architectures can attain low test
error, producing solutions that interpolate accurately, such models  need not
allocate cross-influence across symmetry groups, which  inhibits their ability
to learn the underlying solution operator. 

By explicitly measuring cross-influence between symmetry related states, our framework
reveals whether learning  propagates information coherently along an orbit or instead
distribute influence in a way that precludes convergence to generalizing solutions. This
provides a novel diagnostic for distinguishing models that learn to exploit shared structure
from those that effectively assemble collections of local estimators. Tying symmetry
generalization directly to the geometry of the learned loss landscape complements standard
equivariance accuracy metrics and provides a clearer criterion for when apparent performance
reflects genuine physics learning. Such diagnostics are essential for building trust in
scientific machine learning systems, particularly in applications requiring robustness under
symmetry transformations.  Looking forward, this perspective motivates the development of
approximate or relaxed symmetry mechanisms that retain sufficient structure to guide
generalization while preserving the flexibility needed for efficient optimization.

\section{Acknowledgments} Research presented in this report was
supported by the Laboratory Directed Research and Development program of Los Alamos National Laboratory
under project number(s) 20250637DI, 20250638DI, and
20250639DI. This research used resources provided by the
Los Alamos National Laboratory Institutional Computing
Program, which is supported by the U.S. Department of Energy National Nuclear Security Administration under Contract No. 89233218CNA000001. It is published under 
LA-UR-25-29466.

\bibliography{main}

@misc{equivdyna_2024,
  author  = {Canez, Diego and Midavaine, Nesta and Stessen, Thijs and Fan, Jiapeng and Arias, Sebastian and Garcia, Alejandro},
  title   = {Effect of equivariance on training dynamics},
  journal = {GRaM Workshop, ICML 2024},
  year    = {2024},
  doi     = {10.5281/zenodo.14283519},
  url     = {https://gram-blogposts.github.io/blog/2024/relaxed-equivariance/}
}

@inproceedings{tegan,
author = {Godfrey, Charles and Brown, Davis and Emerson, Tegan and Kvinge, Henry},
booktitle = {Advances in Neural Information Processing Systems},
editor = {S. Koyejo and S. Mohamed and A. Agarwal and D. Belgrave and K. Cho and A. Oh},
pages = {11893--11905},
publisher = {Curran Associates, Inc.},
title = {On the Symmetries of Deep Learning Models and their Internal Representations},
volume = {35},
year = {2022}
}

@misc{wang2022approximatelyequivariantnetworksimperfectly,
      title={Approximately Equivariant Networks for Imperfectly Symmetric Dynamics},
      author={Rui Wang and Robin Walters and Rose Yu},
      year={2022},
      eprint={2201.11969},
      archivePrefix={arXiv},
      primaryClass={cs.LG},
      url={https://arxiv.org/abs/2201.11969},
}

@misc{kayhan2020translationinvariancecnnsconvolutional,
      title={On Translation Invariance in CNNs: Convolutional Layers can Exploit Absolute Spatial Location},
      author={Osman Semih Kayhan and Jan C. van Gemert},
      year={2020},
      eprint={2003.07064},
      archivePrefix={arXiv},
      primaryClass={cs.CV},
      url={https://arxiv.org/abs/2003.07064},
}

@article{JMLR:v20:19-519,
  author  = {Aharon Azulay and Yair Weiss},
  title   = {Why do deep convolutional networks generalize so poorly to small image transformations?},
  journal = {Journal of Machine Learning Research},
  year    = {2019},
  volume  = {20},
  number  = {184},
  pages   = {1--25},
  url     = {http://jmlr.org/papers/v20/19-519.html}
}

@InProceedings{pmlr-v97-zhang19a,
  title = 	 {Making Convolutional Networks Shift-Invariant Again},
  author =       {Zhang, Richard},
  booktitle = 	 {Proceedings of the 36th International Conference on Machine Learning},
  pages = 	 {7324--7334},
  year = 	 {2019},
  editor = 	 {Chaudhuri, Kamalika and Salakhutdinov, Ruslan},
  volume = 	 {97},
  series = 	 {Proceedings of Machine Learning Research},
  month = 	 {09--15 Jun},
  publisher =    {PMLR},
  url = 	 {https://proceedings.mlr.press/v97/zhang19a.html},
}

@misc{zhao2024improvingconvergence,
      title={Improving Convergence and Generalization Using Parameter Symmetries},
      author={Bo Zhao and Robert M. Gower and Robin Walters and Rose Yu},
      year={2024},
      eprint={2305.13404},
      archivePrefix={arXiv},
      primaryClass={cs.LG},
      url={https://arxiv.org/abs/2305.13404},
}

@misc{akhoundsadegh2023liepointsymmetryphysics,
      title={Lie Point Symmetry and Physics Informed Networks},
      author={Tara Akhound-Sadegh and Laurence Perreault-Levasseur and Johannes Brandstetter and Max Welling and Siamak Ravanbakhsh},
      year={2023},
      eprint={2311.04293},
      archivePrefix={arXiv},
      primaryClass={cs.LG},
      url={https://arxiv.org/abs/2311.04293},
}

@misc{gregory2024equivariant,
      title={Equivariant geometric convolutions for emulation of dynamical systems},
      author={Wilson G. Gregory and David W. Hogg and Ben Blum-Smith and Maria Teresa Arias and Kaze W. K. Wong and Soledad Villar},
      year={2024},
      eprint={2305.12585},
      archivePrefix={arXiv},
      primaryClass={cs.LG},
      url={https://arxiv.org/abs/2305.12585},
}

@InProceedings{pmlr-v162-brandstetter22a,
  title = 	 {Lie Point Symmetry Data Augmentation for Neural {PDE} Solvers},
  author =       {Brandstetter, Johannes and Welling, Max and Worrall, Daniel E},
  booktitle = 	 {Proceedings of the 39th International Conference on Machine Learning},
  pages = 	 {2241--2256},
  year = 	 {2022},
  editor = 	 {Chaudhuri, Kamalika and Jegelka, Stefanie and Song, Le and Szepesvari, Csaba and Niu, Gang and Sabato, Sivan},
  volume = 	 {162},
  series = 	 {Proceedings of Machine Learning Research},
  month = 	 {17--23 Jul},
  publisher =    {PMLR},
  url = 	 {https://proceedings.mlr.press/v162/brandstetter22a.html},
}

@misc{fort2020stiffnessnewperspectivegeneralization,
      title={Stiffness: A New Perspective on Generalization in Neural Networks},
      author={Stanislav Fort and Paweł Krzysztof Nowak and Stanislaw Jastrzebski and Srini Narayanan},
      year={2020},
      eprint={1901.09491},
      archivePrefix={arXiv},
      primaryClass={cs.LG},
      url={https://arxiv.org/abs/1901.09491},
}

@misc{arpit2017closerlookmemorizationdeep,
      title={A Closer Look at Memorization in Deep Networks},
      author={Devansh Arpit and Stanislaw Jastrzebski and Nicolas Ballas and David Krueger and Emmanuel Bengio and Maxinder S. Kanwal and Tegan Maharaj and Asja Fischer and Aaron Courville and Yoshua Bengio and Simon Lacoste-Julien},
      year={2017},
      eprint={1706.05394},
      archivePrefix={arXiv},
      primaryClass={stat.ML},
      url={https://arxiv.org/abs/1706.05394},
}

@misc{he2020localelasticityneuralnetworks,
      title={The Local Elasticity of Neural Networks}, 
      author={Hangfeng He and Weijie J. Su},
      year={2020},
      eprint={1910.06943},
      archivePrefix={arXiv},
      primaryClass={cs.LG},
      url={https://arxiv.org/abs/1910.06943}, 
}

@misc{fort2019emergentpropertieslocalgeometry,
      title={Emergent properties of the local geometry of neural loss landscapes}, 
      author={Stanislav Fort and Surya Ganguli},
      year={2019},
      eprint={1910.05929},
      archivePrefix={arXiv},
      primaryClass={cs.LG},
      url={https://arxiv.org/abs/1910.05929}, 
}

@misc{chatterjee2020coherentgradientsapproachunderstanding,
      title={Coherent Gradients: An Approach to Understanding Generalization in Gradient Descent-based Optimization}, 
      author={Satrajit Chatterjee},
      year={2020},
      eprint={2002.10657},
      archivePrefix={arXiv},
      primaryClass={cs.LG},
      url={https://arxiv.org/abs/2002.10657}, 
}

@misc{brandstetter2023messagepassingneuralpde,
      title={Message Passing Neural PDE Solvers},
      author={Johannes Brandstetter and Daniel Worrall and Max Welling},
      year={2023},
      eprint={2202.03376},
      archivePrefix={arXiv},
      primaryClass={cs.LG},
      url={https://arxiv.org/abs/2202.03376},
}

@misc{herde2024poseidonefficientfoundationmodels,
      title={Poseidon: Efficient Foundation Models for PDEs}, 
      author={Maximilian Herde and Bogdan Raonić and Tobias Rohner and Roger Käppeli and Roberto Molinaro and Emmanuel de Bézenac and Siddhartha Mishra},
      year={2024},
      eprint={2405.19101},
      archivePrefix={arXiv},
      primaryClass={cs.LG},
      url={https://arxiv.org/abs/2405.19101}, 
}

@misc{takamoto2024pdebenchextensivebenchmarkscientific,
      title={PDEBENCH: An Extensive Benchmark for Scientific Machine Learning}, 
      author={Makoto Takamoto and Timothy Praditia and Raphael Leiteritz and Dan MacKinlay and Francesco Alesiani and Dirk Pflüger and Mathias Niepert},
      year={2024},
      eprint={2210.07182},
      archivePrefix={arXiv},
      primaryClass={cs.LG},
      url={https://arxiv.org/abs/2210.07182}, 
}

@misc{lippe2023pderefinerachievingaccuratelong,
      title={PDE-Refiner: Achieving Accurate Long Rollouts with Neural PDE Solvers}, 
      author={Phillip Lippe and Bastiaan S. Veeling and Paris Perdikaris and Richard E. Turner and Johannes Brandstetter},
      year={2023},
      eprint={2308.05732},
      archivePrefix={arXiv},
      primaryClass={cs.LG},
      url={https://arxiv.org/abs/2308.05732}, 
}

@misc{gupta2022multispatiotemporalscalegeneralizedpdemodeling,
      title={Towards Multi-spatiotemporal-scale Generalized PDE Modeling}, 
      author={Jayesh K. Gupta and Johannes Brandstetter},
      year={2022},
      eprint={2209.15616},
      archivePrefix={arXiv},
      primaryClass={cs.LG},
      url={https://arxiv.org/abs/2209.15616}, 
}

@misc{ohana2025welllargescalecollectiondiverse,
      title={The Well: a Large-Scale Collection of Diverse Physics Simulations for Machine Learning}, 
      author={Ruben Ohana and Michael McCabe and Lucas Meyer and Rudy Morel and Fruzsina J. Agocs and Miguel Beneitez and Marsha Berger and Blakesley Burkhart and Keaton Burns and Stuart B. Dalziel and Drummond B. Fielding and Daniel Fortunato and Jared A. Goldberg and Keiya Hirashima and Yan-Fei Jiang and Rich R. Kerswell and Suryanarayana Maddu and Jonah Miller and Payel Mukhopadhyay and Stefan S. Nixon and Jeff Shen and Romain Watteaux and Bruno Régaldo-Saint Blancard and François Rozet and Liam H. Parker and Miles Cranmer and Shirley Ho},
      year={2025},
      eprint={2412.00568},
      archivePrefix={arXiv},
      primaryClass={cs.LG},
      url={https://arxiv.org/abs/2412.00568}, 
}

@misc{Li2021FNO,
      title={Fourier Neural Operator for Parametric Partial Differential Equations},
      author={Zongyi Li and Nikola Kovachki and Kamyar Azizzadenesheli and Burigede Liu and Kaushik Bhattacharya and Andrew Stuart and Anima Anandkumar},
      year={2021},
      eprint={2010.08895},
      archivePrefix={arXiv},
      primaryClass={cs.LG},
      url={https://arxiv.org/abs/2010.08895},
}

@misc{liu2021swintransformerhierarchicalvision,
      title={Swin Transformer: Hierarchical Vision Transformer using Shifted Windows},
      author={Ze Liu and Yutong Lin and Yue Cao and Han Hu and Yixuan Wei and Zheng Zhang and Stephen Lin and Baining Guo},
      year={2021},
      eprint={2103.14030},
      archivePrefix={arXiv},
      primaryClass={cs.CV},
      url={https://arxiv.org/abs/2103.14030},
}

@article{Lu2021DeepONet,
   title={Learning nonlinear operators via DeepONet based on the universal approximation theorem of operators},
   volume={3},
   ISSN={2522-5839},
   url={http://dx.doi.org/10.1038/s42256-021-00302-5},
   DOI={10.1038/s42256-021-00302-5},
   number={3},
   journal={Nature Machine Intelligence},
   publisher={Springer Science and Business Media LLC},
   author={Lu, Lu and Jin, Pengzhan and Pang, Guofei and Zhang, Zhongqiang and Karniadakis, George Em},
   year={2021},
   pages={218–229}
}

@misc{koh2020understandingblackboxpredictionsinfluence,
      title={Understanding Black-box Predictions via Influence Functions},
      author={Pang Wei Koh and Percy Liang},
      year={2020},
      eprint={1703.04730},
      archivePrefix={arXiv},
      primaryClass={stat.ML},
      url={https://arxiv.org/abs/1703.04730},
}

@book{huber2009robust,
  title        = {Robust Statistics},
  author       = {Huber, Peter J. and Ronchetti, Elvezio M.},
  series       = {Wiley Series in Probability and Statistics},
  publisher    = {John Wiley \& Sons, Inc.},
  year         = {2009},
  isbn         = {9780470129906},
  doi          = {10.1002/9780470434697},
  url          = {https://doi.org/10.1002/9780470434697},
}

@software{TransferLab_team_pyDVL_2024,
author = {TransferLab},
doi = {10.5281/zenodo.10966754},
title = {{pyDVL}},
url = {https://github.com/aai-institute/pyDVL},
version = {v0.9.0},
year = {2024}
}

@software{george_nngeometry,
  author       = {Thomas George},
  title        = {{NNGeometry: Easy and Fast Fisher Information
                   Matrices and Neural Tangent Kernels in PyTorch}},
  year         = {2021},
  publisher    = {Zenodo},
  version      = {v0.3},
  doi          = {10.5281/zenodo.4532597},
  url          = {https://doi.org/10.5281/zenodo.4532597}
}

@article{montoison-orban-2023,
  author  = {Montoison, Alexis and Orban, Dominique},
  title   = {{Krylov.jl: A Julia basket of hand-picked Krylov methods}},
  journal = {Journal of Open Source Software},
  volume  = {8},
  number  = {89},
  pages   = {5187},
  year    = {2023},
  doi     = {10.21105/joss.05187}
}

@book{craig,
author = {Orban, Dominique and Arioli, Mario},
title = {Iterative Solution of Symmetric Quasi-Definite Linear Systems},
publisher = {Society for Industrial and Applied Mathematics},
year = {2017},
doi = {10.1137/1.9781611974737},
address = {Philadelphia, PA},
edition   = {},
URL = {https://epubs.siam.org/doi/abs/10.1137/1.9781611974737},
}

@misc{jacot2020neuraltangentkernelconvergence,
      title={Neural Tangent Kernel: Convergence and Generalization in Neural Networks},
      author={Arthur Jacot and Franck Gabriel and Clément Hongler},
      year={2020},
      eprint={1806.07572},
      archivePrefix={arXiv},
      primaryClass={cs.LG},
      url={https://arxiv.org/abs/1806.07572},
}

@article{JMLR:v21:17-678,
  author  = {James Martens},
  title   = {New Insights and Perspectives on the Natural Gradient Method},
  journal = {Journal of Machine Learning Research},
  year    = {2020},
  volume  = {21},
  number  = {146},
  pages   = {1--76},
  url     = {http://jmlr.org/papers/v21/17-678.html}
}

@misc{kingma2017adammethodstochasticoptimization,
      title={Adam: A Method for Stochastic Optimization}, 
      author={Diederik P. Kingma and Jimmy Ba},
      year={2017},
      eprint={1412.6980},
      archivePrefix={arXiv},
      primaryClass={cs.LG},
      url={https://arxiv.org/abs/1412.6980}, 
}

@software{pal2023lux,
  author    = {Pal, Avik},
  title     = {{Lux: Explicit Parameterization of Deep Neural Networks in Julia}},
  year      = {2023},
  publisher = {Zenodo},
  version   = {v1.4.2},
  doi       = {10.5281/zenodo.7808903},
  url       = {https://doi.org/10.5281/zenodo.7808903},
}

@thesis{pal2023efficient,
  title     = {{On Efficient Training \& Inference of Neural Differential Equations}},
  author    = {Pal, Avik},
  year      = {2023},
  school    = {Massachusetts Institute of Technology}
}

@article{DanischKrumbiegel2021,
  doi = {10.21105/joss.03349},
  url = {https://doi.org/10.21105/joss.03349},
  year = {2021},
  publisher = {The Open Journal},
  volume = {6},
  number = {65},
  pages = {3349},
  author = {Simon Danisch and Julius Krumbiegel},
  title = {{Makie.jl}: Flexible high-performance data visualization for {Julia}},
  journal = {Journal of Open Source Software}
}

@article{Zygote.jl-2018,
  author    = {Michael Innes},
  title     = {Don't Unroll Adjoint: Differentiating SSA-Form Programs},
  journal   = {CoRR},
  volume    = {abs/1810.07951},
  year      = {2018},
  url       = {http://arxiv.org/abs/1810.07951},
  archivePrefix = {arXiv},
  eprint    = {1810.07951},
}

@inproceedings{ye2024pdeformer,
    title={{PDE}former: Towards a Foundation Model for One-Dimensional Partial Differential Equations},
    author={Zhanhong Ye and Xiang Huang and Leheng Chen and Hongsheng Liu and Zidong Wang and Bin Dong},
    booktitle={ICLR 2024 Workshop on AI4DifferentialEquations In Science},
    year={2024},
    url={https://openreview.net/forum?id=GLDMCwdhTK}
}

@misc{subramanian2023foundationmodelsscientificmachine,
      title={Towards Foundation Models for Scientific Machine Learning: Characterizing Scaling and Transfer Behavior}, 
      author={Shashank Subramanian and Peter Harrington and Kurt Keutzer and Wahid Bhimji and Dmitriy Morozov and Michael Mahoney and Amir Gholami},
      year={2023},
      eprint={2306.00258},
      archivePrefix={arXiv},
      primaryClass={cs.LG},
      url={https://arxiv.org/abs/2306.00258}, 
}

@misc{gradpinn,
      title={Gradient Alignment in Physics-informed Neural Networks: A Second-Order Optimization Perspective}, 
      author={Sifan Wang and Ananyae Kumar Bhartari and Bowen Li and Paris Perdikaris},
      year={2025},
      eprint={2502.00604},
      archivePrefix={arXiv},
      primaryClass={cs.LG},
      url={https://arxiv.org/abs/2502.00604}, 
}

@inproceedings{CohenWelling2016,
  title={Group Equivariant Convolutional Networks},
  author={Cohen, Taco S. and Welling, Max},
  booktitle={Proceedings of the 33rd International Conference on Machine Learning},
  series={PMLR},
  volume={48},
  pages={2990--2999},
  year={2016},
  url={https://proceedings.mlr.press/v48/cohenc16.pdf}
}

@inproceedings{CohenWelling2017,
  title={Steerable {CNN}s},
  author={Cohen, Taco S. and Welling, Max},
  booktitle={ICLR},
  year={2017},
  url={https://arxiv.org/abs/1612.08498}
}

@article{Weiler2018,
  title={3D Steerable {CNN}s: Learning Rotationally Equivariant Features in Volumetric Data},
  author={Weiler, Maurice and Geiger, Mario and Welling, Max and Boomsma, Wouter and Cohen, Taco},
  journal={arXiv:1807.02547},
  year={2018},
  url={https://arxiv.org/abs/1807.02547}
}

@article{Thomas2018,
  title={Tensor Field Networks: Rotation- and Translation-Equivariant Neural Networks for 3D Point Clouds},
  author={Thomas, Nathaniel and Smidt, Tess and Kearnes, Steven and Yang, Lusann and Li, Li and Kohlhoff, Kai and Riley, Patrick},
  journal={arXiv:1802.08219},
  year={2018},
  url={https://arxiv.org/abs/1802.08219}
}

@inproceedings{Satorras2021,
  title={E(n) Equivariant Graph Neural Networks},
  author={Garcia Satorras, Victor and Hoogeboom, Emiel and Welling, Max},
  booktitle={Proceedings of the 38th International Conference on Machine Learning},
  series={PMLR},
  volume={139},
  pages={9323--9332},
  year={2021},
  url={https://proceedings.mlr.press/v139/satorras21a.html}
}

@article{Gruver2022,
  title={The Lie Derivative for Measuring Learned Equivariance},
  author={Gruver, Nate and Finzi, Marc and Goldblum, Micah and Wilson, Andrew Gordon},
  journal={arXiv:2210.02984},
  year={2022},
  url={https://arxiv.org/abs/2210.02984}
}

@inproceedings{BasuEtAl2020,
  title={On Second-Order Group Influence Functions for Black-Box Predictions},
  author={Basu, Samyadeep and You, Xuchen and Feizi, Soheil},
  booktitle={Proceedings of the 37th International Conference on Machine Learning},
  series={PMLR},
  volume={119},
  pages={715--724},
  year={2020},
  url={https://proceedings.mlr.press/v119/basu20b.html}
}

@misc{finzi2021,
      title={Residual Pathway Priors for Soft Equivariance Constraints}, 
      author={Marc Finzi and Gregory Benton and Andrew Gordon Wilson},
      year={2021},
      eprint={2112.01388},
      archivePrefix={arXiv},
      primaryClass={cs.LG},
      url={https://arxiv.org/abs/2112.01388}, 
}

@misc{climax,
      title={ClimaX: A foundation model for weather and climate}, 
      author={Tung Nguyen and Johannes Brandstetter and Ashish Kapoor and Jayesh K. Gupta and Aditya Grover},
      year={2023},
      eprint={2301.10343},
      archivePrefix={arXiv},
      primaryClass={cs.LG},
      url={https://arxiv.org/abs/2301.10343}, 
}

@misc{aurora,
      title={A Foundation Model for the Earth System}, 
      author={Cristian Bodnar and Wessel P. Bruinsma and Ana Lucic and Megan Stanley and Anna Vaughan and Johannes Brandstetter and Patrick Garvan and Maik Riechert and Jonathan A. Weyn and Haiyu Dong and Jayesh K. Gupta and Kit Thambiratnam and Alexander T. Archibald and Chun-Chieh Wu and Elizabeth Heider and Max Welling and Richard E. Turner and Paris Perdikaris},
      year={2024},
      eprint={2405.13063},
      archivePrefix={arXiv},
      primaryClass={physics.ao-ph},
      url={https://arxiv.org/abs/2405.13063}, 
}

@misc{xie2025talesymmetriesexploringloss,
      title={A Tale of Two Symmetries: Exploring the Loss Landscape of Equivariant Models}, 
      author={YuQing Xie and Tess Smidt},
      year={2025},
      eprint={2506.02269},
      archivePrefix={arXiv},
      primaryClass={cs.LG},
      url={https://arxiv.org/abs/2506.02269}, 
}

@book{absil2008,
  title     = {Optimization Algorithms on Matrix Manifolds},
  author    = {Absil, P.-A. and Mahony, R. and Sepulchre, R.},
  publisher = {Princeton University Press},
  address   = {Princeton, NJ},
  year      = {2008},
  isbn      = {978-0-691-13298-3}
}

@article{lu1997standardized,
  author  = {Lu, Jiandong and Ko, Daijin and Chang, Ted},
  title   = {The Standardized Influence Matrix and Its Applications},
  journal = {Journal of the American Statistical Association},
  volume  = {92},
  number  = {440},
  pages   = {1572--1580},
  year    = {1997},
  issn    = {0162-1459},
  url     = {https://www.jstor.org/stable/2965428},
}

@article{heritier1994robust,
  author  = {H{\'e}ritier, St{\'e}phane and Ronchetti, Elvezio},
  title   = {Robust Bounded-Influence Tests in General Parametric Models},
  journal = {Journal of the American Statistical Association},
  volume  = {89},
  number  = {427},
  pages   = {897--904},
  year    = {1994},
  issn    = {0162-1459},
  url     = {https://www.jstor.org/stable/2290914},

}

@misc{rautela2026morphpdefoundationmodels,
      title={MORPH: PDE Foundation Models with Arbitrary Data Modality}, 
      author={Mahindra Singh Rautela and Alexander Most and Siddharth Mansingh and Bradley C. Love and Alexander Scheinker and Diane Oyen and Nathan Debardeleben and Earl Lawrence and Ayan Biswas},
      year={2026},
      eprint={2509.21670},
      archivePrefix={arXiv},
      primaryClass={cs.CV},
      url={https://arxiv.org/abs/2509.21670}, 
}

@misc{rigno,
  title={RIGNO: A Graph-based framework for robust and accurate operator learning for PDEs on arbitrary domains},
  author={Sepehr Mousavi and Shizheng Wen and Levi Lingsch and Maximilian Herde and Bogdan Raonić and Siddhartha Mishra},
  year={2025},
  eprint={2501.19205},
  archivePrefix={arXiv},
  primaryClass={cs.LG},
  url={https://arxiv.org/abs/2501.19205},
}

@misc{walrus,
  title={Walrus: A Cross-Domain Foundation Model for Continuum Dynamics},
  author={Michael McCabe and Payel Mukhopadhyay and Tanya Marwah and Bruno Regaldo-Saint Blancard and Francois Rozet and Cristiana Diaconu and Lucas Meyer and Kaze W. K. Wong and Hadi Sotoudeh and Alberto Bietti and Irina Espejo and Rio Fear and Siavash Golkar and Tom Hehir and Keiya Hirashima and Geraud Krawezik and Francois Lanusse and Rudy Morel and Ruben Ohana and Liam Parker and Mariel Pettee and Jeff Shen and Kyunghyun Cho and Miles Cranmer and Shirley Ho},
  year={2026},
  eprint={2511.15684},
  archivePrefix={arXiv},
  primaryClass={cs.LG},
  url={https://arxiv.org/abs/2511.15684},
}

@misc{paper1,
  title={Generalization vs. Memorization in Autoregressive Deep Learning: Or, Examining Temporal Decay of Gradient Coherence},
  author={James Amarel and Nicolas Hengartner and Robyn Miller and Kamaljeet Singh and Siddharth Mansingh and Arvind Mohan and Benjamin Migliori and Emily Casleton and Alexei Skurikhin and Earl Lawrence and Gerd J. Kunde},
  year={2026},
  eprint={2509.00024},
  archivePrefix={arXiv},
  primaryClass={physics.comp-ph},
  url={https://arxiv.org/abs/2509.00024},
}

\appendix
\section*{Software and Data}
The code used in this work is available at 
\href{https://github.com/lanl/PDEHats}{https://github.com/lanl/PDEHats}.
We trained our models on
the openly available dataset PDEGym
\citep{herde2024poseidonefficientfoundationmodels}   
using Lux.jl
\citep{pal2023lux, pal2023efficient}, with Zygote.jl as our auto-differentiation
backend \citep{Zygote.jl-2018}. Plots in this manuscript were generated using
Makie.jl \citep{DanischKrumbiegel2021}.

\section{Data Selection.}
\label{app:data}
PDEGym data is particularly valuable for studying the progression from a linear wave regime
with discontinuities to fully developed turbulence, a crossover that poses computational and
analytical challenges due to the presence of sharp wave-fronts and emergent nonlinear
interactions. While the CE flows exhibit comparable large-scale structures, they also
display qualitatively distinct behaviors. In particular, CE-RPUI initial configurations give
rise to complex finger-like instabilities in the flow field that are absent or less
pronounced in both CE-RP and CE-CRP.  In total, we used \(6,500\) trajectories  for each of
the \(3\) classes of initial conditions; for each trajectory, we used the first \(16\) time
steps, for total of \(104,000\) training pairs per initial condition class, requiring greater than \(150\)
GB memory. 

\section{Model Selection.}
\label{app:model}
We focus on UNet and ViT backbones because they represent the dominant scalable
image-to-image paradigms currently used in neural PDE emulation and PDE foundation modeling.
Convolutional backbones derived from UNets remain standard architectures for grid-based
surrogate modeling and large-scale benchmarking efforts
\citep{takamoto2024pdebenchextensivebenchmarkscientific,
gupta2022multispatiotemporalscalegeneralizedpdemodeling}. Likewise, transformer-based
backbones now underpin many recent PDE foundation models and large-scale operator learners,
including Poseidon and PDEformer \citep{herde2024poseidonefficientfoundationmodels,
ye2024pdeformer}, in addition to several contemporary multi-physics
\citep{rautela2026morphpdefoundationmodels} and shifting window transformer 
\citep{liu2021swintransformerhierarchicalvision} architectures. Fourier neural operators
provide an important complementary class of operator-learning models \citep{Li2021FNO,
subramanian2023foundationmodelsscientificmachine}; however, because their representations
are constructed from truncated Fourier modes, the resulting plane-wave basis is less
naturally suited to the spatially localized discontinuities and sharp shock fronts
characteristic of compressible flow evolution
\citep{ohana2025welllargescalecollectiondiverse}. Consistent with this expectation, we found
FNO variants to be noncompetitive with both UNet and ViT backbones on the flows considered
in this work, which is consistent with recent results by the PDEGym authors \citep{rigno}.
DeepONets provide a foundational operator-learning formalism \citep{Lu2021DeepONet}, but are
not the prevailing architecture for contemporary large-scale PDE solution operator
emulation; this assertion is consistent with the present lack of competitive DeepONet models
on PDEGym.

\section{Loss Sensitivity as Gradient Overlap}
\label{app:inf}
The influence diagnostic admits a simple interpretation as a first-order sensitivity of the loss function. Let \(C_p\)
denote the cost associated with a perturbing example \(p\), and let the induced
parameter displacement be
    \(\delta \theta_p = - \eta^{-1} \nabla_\theta C_p\),
where \(\eta\) is the Gauss-Newton metric. The corresponding first-order change
in the cost of a response example \(r\) is
    \(C_r(\theta+\delta\theta_p)-C_r(\theta)
    \approx
    -\left\langle
    \nabla_\theta C_r,
    \eta^{-1}\nabla_\theta C_p
    \right\rangle\).
Thus, gradient alignment
measures the first-order sensitivity of example \(r\) to an update generated by
example \(p\). Positive alignment indicates that the perturbing update acts
coherently on the response example, whereas weak or negative alignment indicates
that the corresponding learning signals are decoupled or conflicting. 

\section{Determination of $\eta^{-1}$}
\label{app:qr}

Influence-function computations in deep learning are commonly performed either
through iterative matrix-free solvers, such as conjugate-gradient methods
\citep{craig, montoison-orban-2023}, or through scalable approximations to the
curvature operator \citep{george_nngeometry, TransferLab_team_pyDVL_2024}.
While such approaches are often necessary in high-dimensional parameter spaces,
iterative methods require repeated Hessian-vector products and convergence
tolerances, whereas low-rank or diagonal approximations generically do not
provide controlled error estimates.

In the present setting, however, the influence observable depends only on
metric-weighted overlaps between example gradients. Let \(G^n_{\alpha} =
\partial_{\alpha} C^n\) denote the gradient associated with mini-batch example
\(n\). Performing a QR decomposition on \(G\) yields an orthonormal basis \(Q\)
that projects onto the gradient subspace. Hence, the Gauss-Newton metric
can be efficiently
inverted. Since the dimension of the gradient-spanned subspace is bounded by
the mini-batch size rather than the total number of trainable parameters, the
resulting inversion is both exact and computationally efficient, eliminating
the need for uncontrolled curvature approximations or iterative parameter-space
solves.

For CE and NS data, hat-matrix elements were determined for \(18\)
different mini-batches, each of which contained \(3\) trajectories 
corresponding to distinct initial conditions over \(16\) time-steps, for each
seed of each model architecture trained. In practice, we additionally include
an \(\ell_2\)-type regularizer corresponding to the weight-decay term used
during AdamW optimization
\citep{kingma2017adammethodstochasticoptimization}, \(\eta_{\mu\nu}
\rightarrow
\eta_{\mu\nu}
+
\lambda \delta_{\mu\nu}\),
with \(\lambda\) weakly lifting the zero modes of \(\eta\).
Our results are not sensitive to choosing either \(10^{-6}\) or \(10^{-4}\) for
\(\lambda\) strength.

\section{Limitations}
\label{app:limits}
Certain physically relevant symmetries remain outside the scope of the present study.
Galilean boosts, in particular, impose a highly restrictive constraint: exact equivariance
can be achieved only by affine transformations, while generic nonlinear architectures can at
best satisfy this symmetry approximately
\citep{wang2022approximatelyequivariantnetworksimperfectly}. It would be interesting to
extend our influence-based analysis to probe the response under small Galilean boosts. We
did not examine the scaling symmetry of the continuum Navier-Stokes equations. In practical
settings this symmetry is generically broken by spatial discretization, numerical
regularization, and coarse graining of the data, rendering its faithful assessment ambiguous
within the present framework. 
Finally, our analysis is restricted to two widely used architectural backbones, namely UNets
and Vision Transformers. While these models represent dominant paradigms in contemporary
large-scale PDE emulation, they do not exhaust the space of operator-learning methods.
Consequently, the degree to which our measured gradient coherence properties depend on the
chosen backbone rather than on the autoregressive training paradigm itself remains only
partially resolved. 

Finally, the reported influence profiles aggregate over multiple trajectory
times and initial-condition classes, so their variability reflects not only
estimator noise but also the intrinsic heterogeneity of the operator-learning
problem: different flow regimes present distinct local loss geometries, and
symmetry-compatible gradient transport need not be equally expressed across
shocks, smooth advective structures, and vorticity-dominated states.

\section{Supplementary Figures}

\begin{figure}[htbp]
\floatconts
  {fig:ZV_CE}
  {\caption{
Diagnostics under vertical translations on CE data.
Markers and rangebars indicate the median and interquartile range over seeds.
Nonuniform dependence on translation distance reflects vertical symmetry
breaking and complements the horizontal response in \autoref{fig:ZH_CE}.
  }}
  {%
    \setlength{\jmlrminsubcaptionwidth}{0.40\linewidth}%
    \subfigure[
Relative equivariance error as a function of vertical translation.
    ][t]{%
      \includegraphics[width=0.40\linewidth]{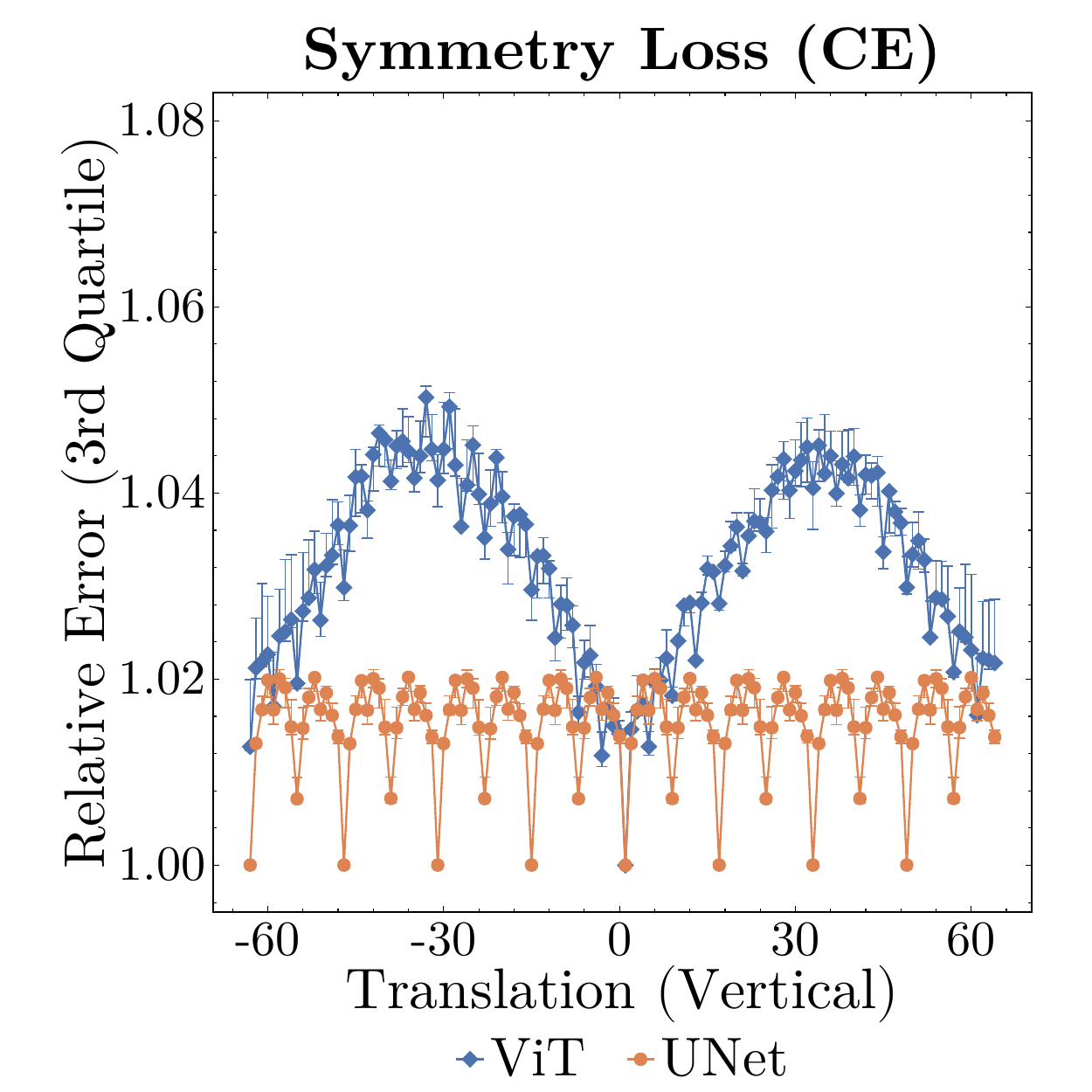}}%
    \hfill
    \subfigure[
Gradient alignment between an example and its vertically translated state.
    ][t]{%
      \includegraphics[width=0.40\linewidth]{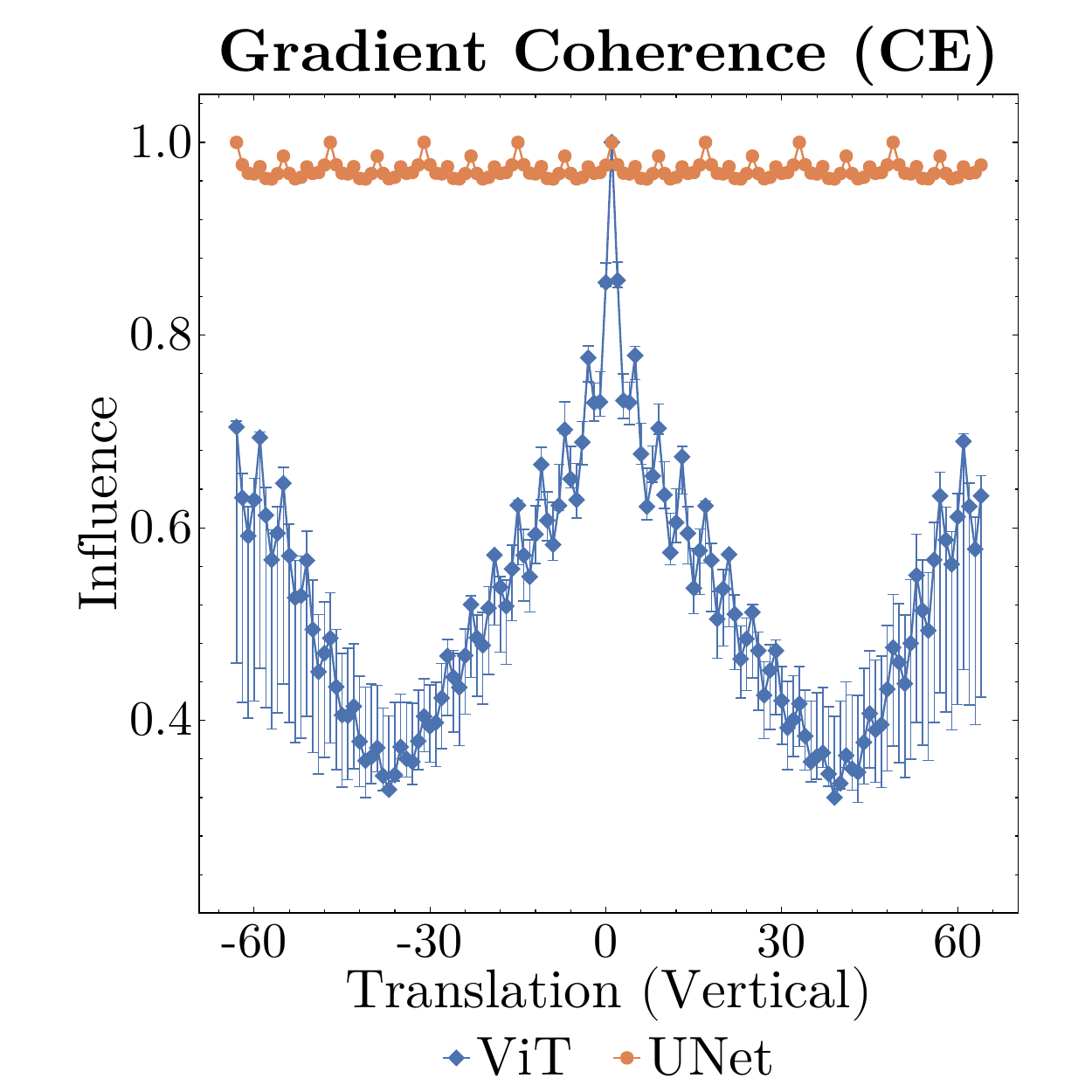}}
  }%
\end{figure}

\begin{figure}[htbp]
\floatconts
  {fig:ZV_NS}
  {\caption{
Diagnostics under vertical translations on Navier-Stokes (NS) data.
Markers and rangebars indicate the median and interquartile range over seeds.
Nonuniform dependence on translation distance reflects vertical symmetry breaking;
for horizontal translations, see \autoref{fig:ZH_NS}.
  }}
  {%
    \setlength{\jmlrminsubcaptionwidth}{0.40\linewidth}%
    \subfigure[
Relative equivariance error as a function of vertical translation.
    ][t]{%
      \includegraphics[width=0.40\linewidth]{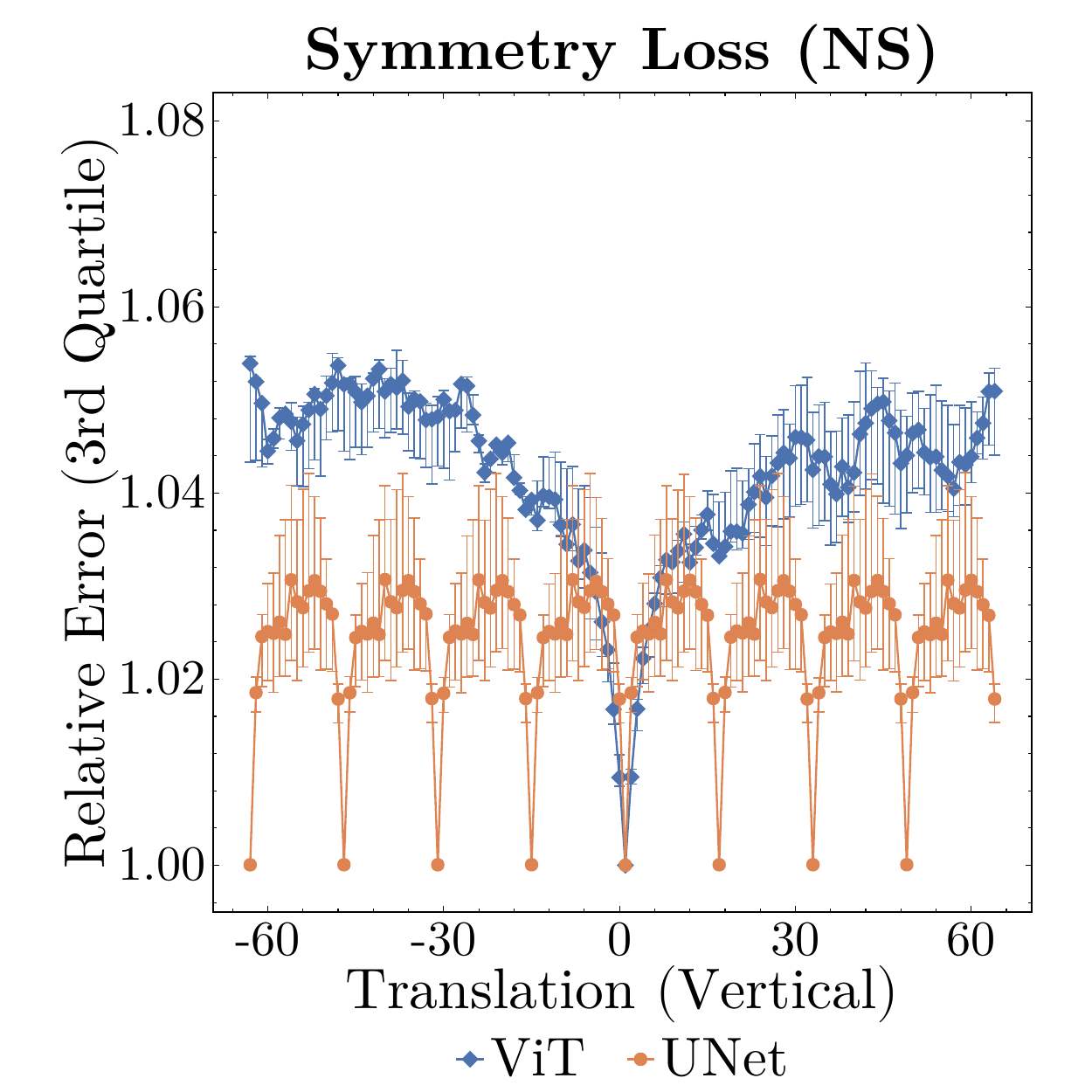}}%
    \hfill
    \subfigure[
Gradient alignment between an example and its vertically translated state.
    ][t]{%
      \includegraphics[width=0.40\linewidth]{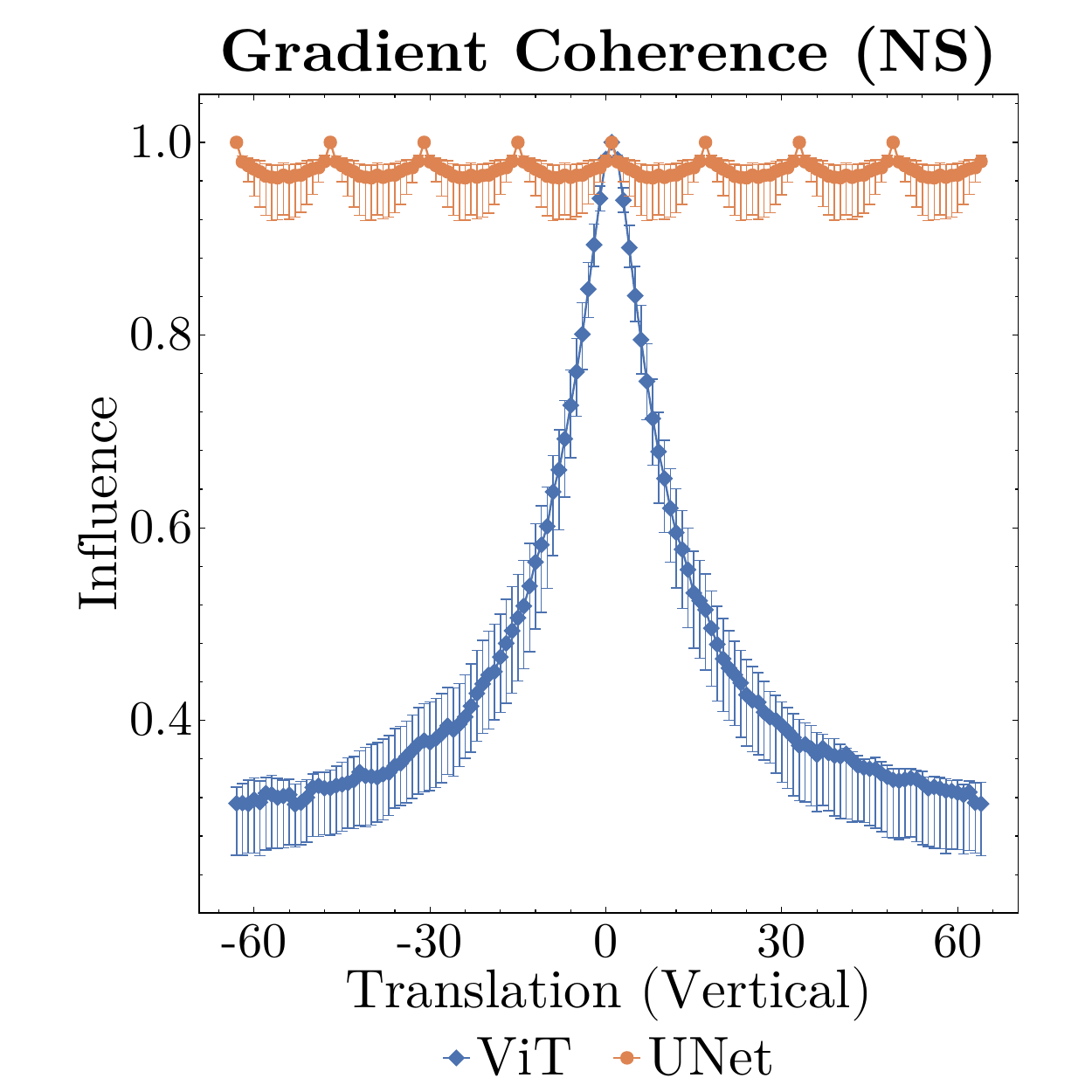}}
  }%
\end{figure}

\begin{figure}[htbp]
\floatconts
  {fig:ZH_NS}
  {\caption{
Diagnostics under horizontal translations on Navier-Stokes (NS) data.
Markers and rangebars indicate the median and interquartile range over seeds.
Nonuniform dependence on translation distance reflects horizontal symmetry breaking;
for vertical translations, see \autoref{fig:ZV_NS}.
  }}
  {%
    \setlength{\jmlrminsubcaptionwidth}{0.40\linewidth}%
    \subfigure[
Relative equivariance error as a function of horizontal translation.
    ][t]{%
      \includegraphics[width=0.40\linewidth]{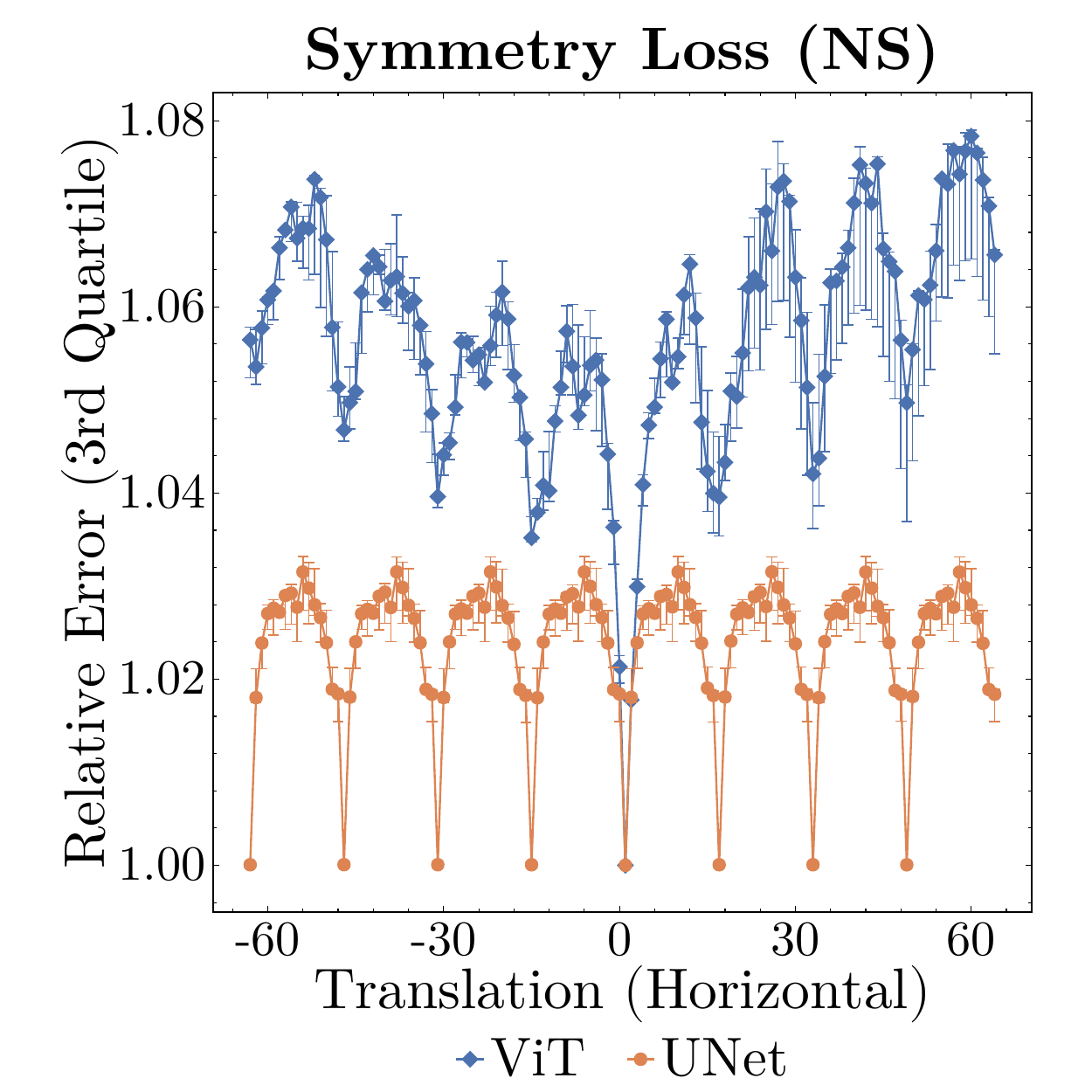}}%
    \hfill
    \subfigure[
Gradient alignment between an example and its horizontally translated state.
    ][t]{%
      \includegraphics[width=0.40\linewidth]{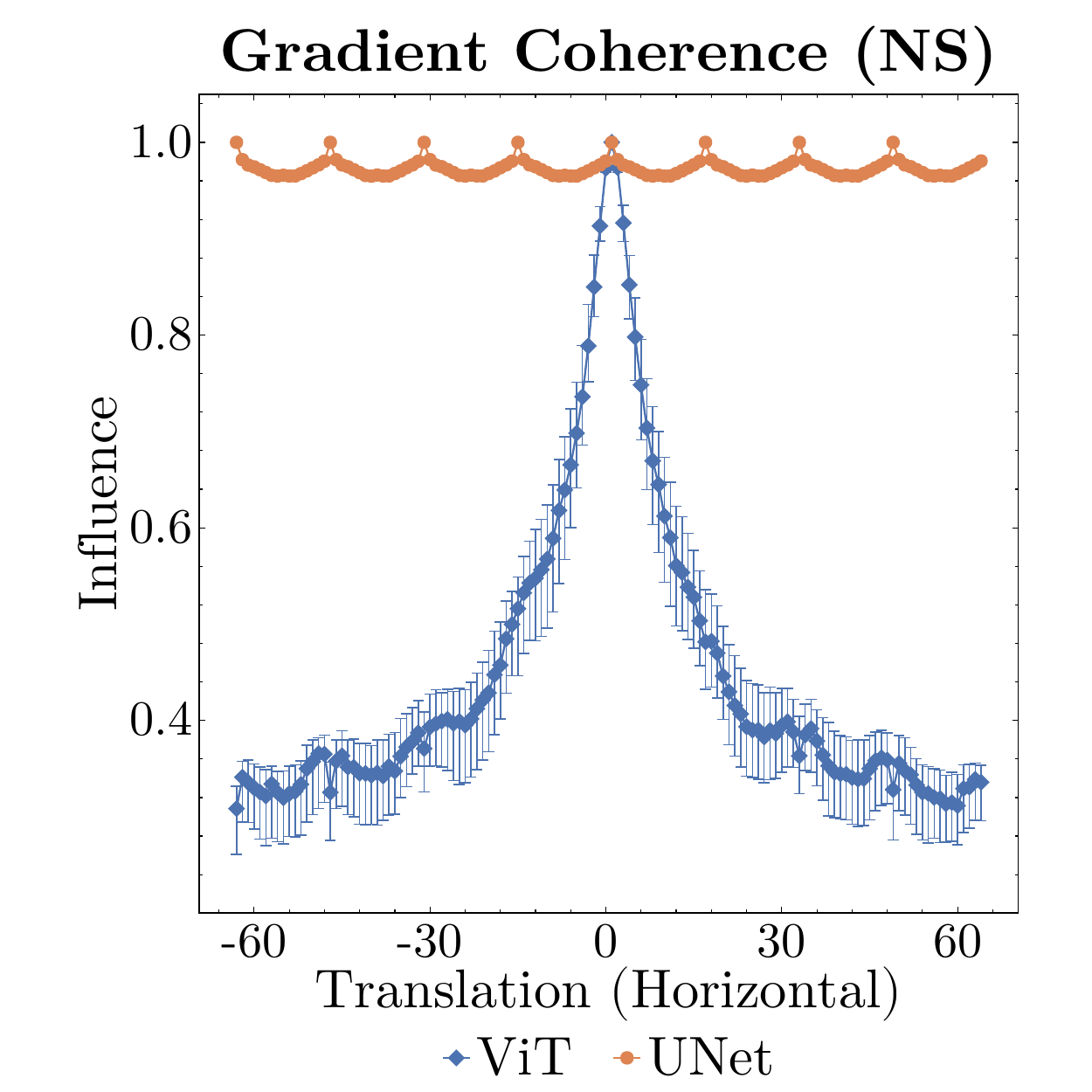}}
  }%
\end{figure}

\begin{figure}[htbp]
\floatconts
  {fig:D4_CE}
  {\caption{
Diagnostics under \(D_{4}\) symmetry on CE data.
Markers and rangebars indicate the median and interquartile range over seeds.
This error-wise uniformity is consistent with the broadly uniform cross-influence
profile, indicating that gradient updates couple across the sampled
\(D_{4}\) actions. The CE case therefore contrasts with the NS diagnostics in
\autoref{fig:D4_NS}, where equivariance error and influence vary strongly across
group elements.
Since our models are evaluated in the late-stage learning regime [see
\autoref{fig:opt_curves}], the observed off-identity gradient alignment
supports continued symmetry-coherent gradient transfer across the \(D_4\)
orbit.
Pearson correlations, computed over all measurements across batches, time
steps, initial-condition classes, and \(D_4\) elements, except the identity, between relative
equivariance error and cross-influence normalized by self-influence on the
identity element are \(-0.17\pm 0.01\) for both UNet and ViT,
where \(\pm\) denotes the standard deviation about the mean over model seeds.
  }}
  {%
    \setlength{\jmlrminsubcaptionwidth}{0.40\linewidth}%
    \subfigure[
Relative equivariance error for each \(D_{4}\) group action.
    ][t]{%
      \includegraphics[width=0.40\linewidth]{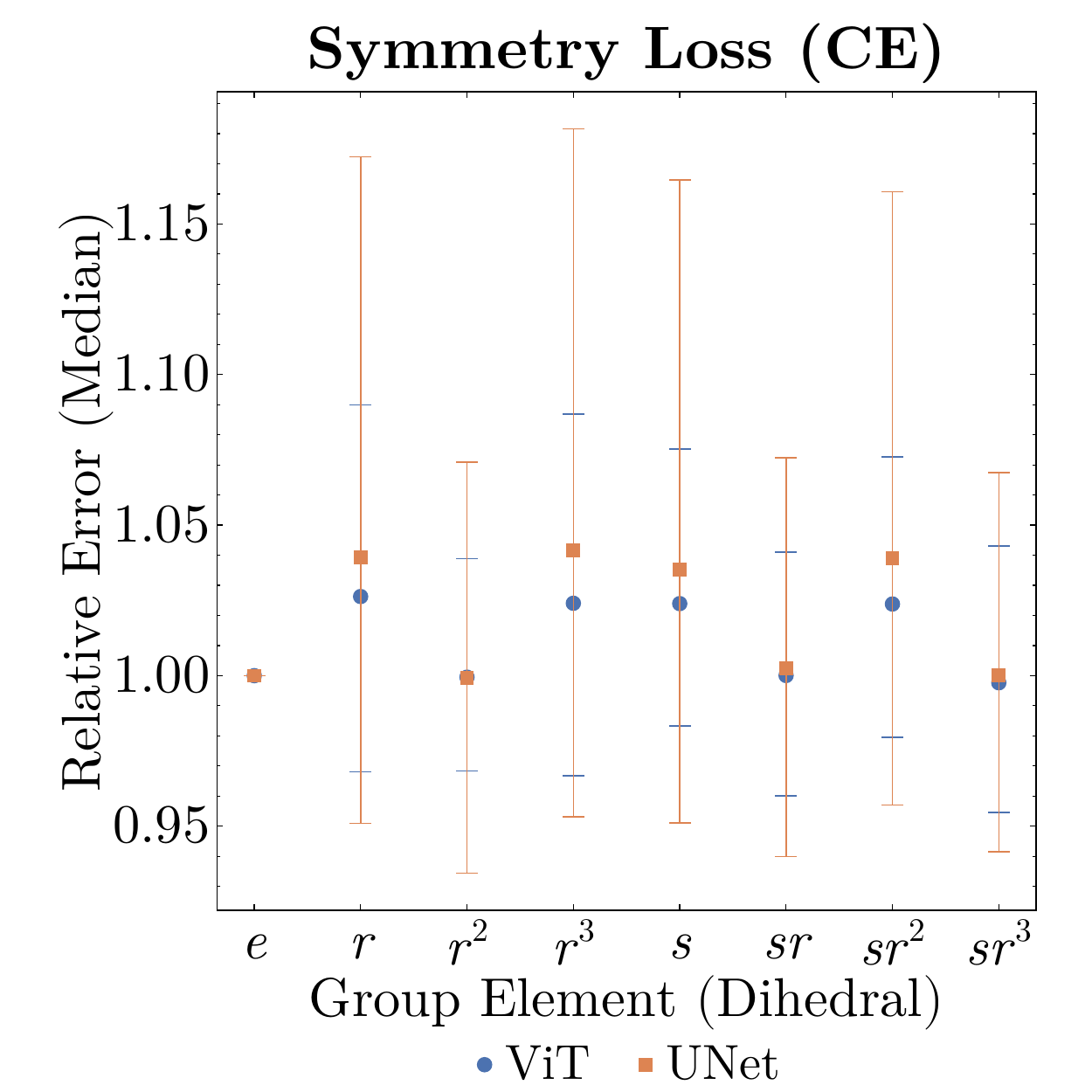}}%
    \hfill
    \subfigure[
Gradient alignment between an example and its \(D_{4}\)-transformed state.
    ][t]{%
      \includegraphics[width=0.40\linewidth]{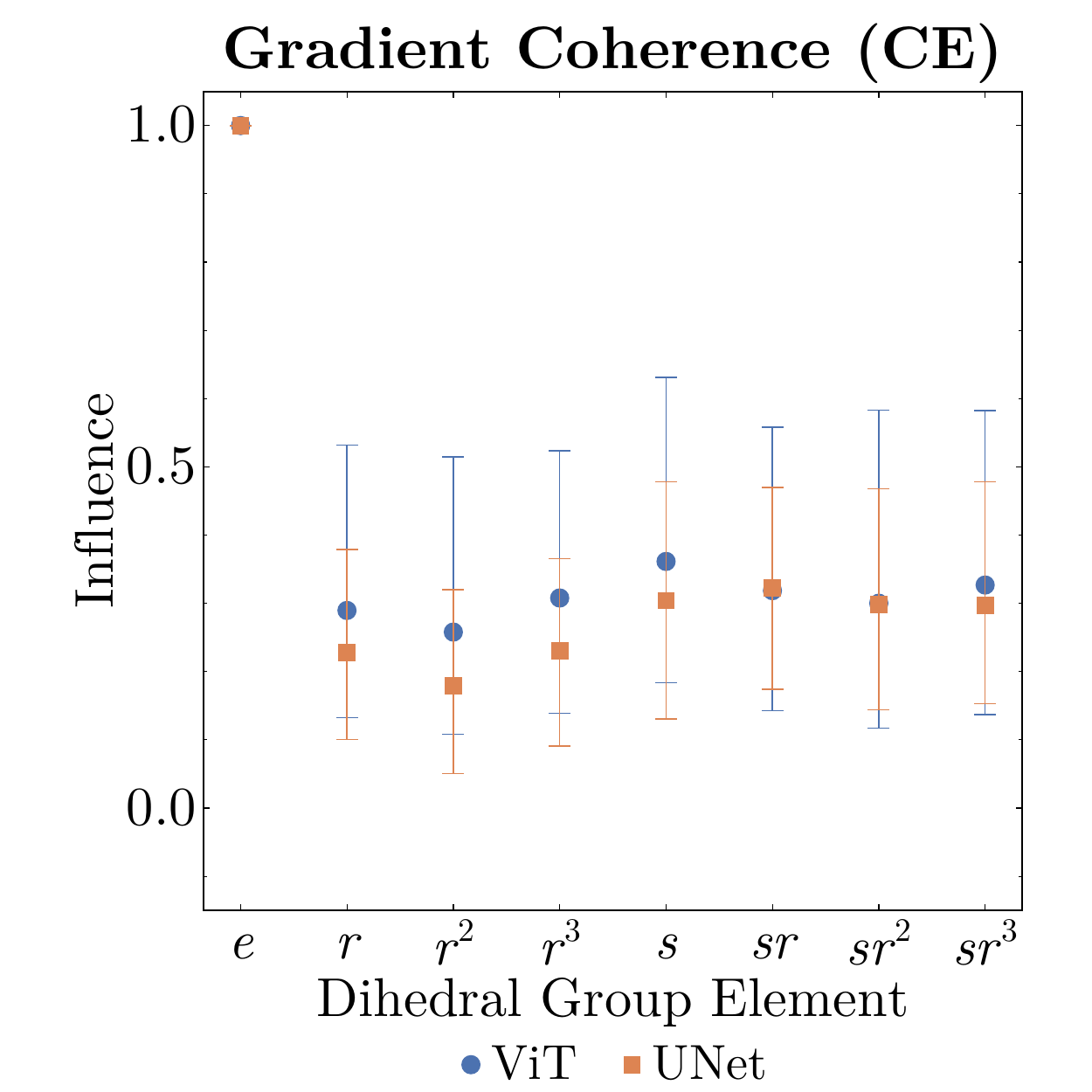}}
  }%
\end{figure}

\begin{figure}[htbp]
\floatconts
  {fig:Klein_NS}
  {\caption{
    Evolution of cross-influence within the residual Klein four-subgroup
    \(V_{4}=\{e,r^{2},sr,sr^{3}\}\) and its complement in \(D_{4}\) for
    Navier--Stokes models. Influence is aggregated over the nonidentity learned
    transformations \(V_{4}\setminus e\) and the unlearned transformations
    \(D_{4}\setminus V_{4}\). The two classes exhibit comparable cross-influence
    at initialization, but separate rapidly during training: transformations in
    \(D_{4}\setminus V_{4}\) receive systematically less cross-influence than
    those in \(V_{4}\setminus e\) across subsequent epochs. This separation
    persists for both architectures. Markers and rangebars indicate the median
    and interquartile range over seeds.
  }}
  {%
    \setlength{\jmlrminsubcaptionwidth}{0.40\linewidth}%
    \subfigure[
      Cross-influence into the learned and unlearned \(D_{4}\) transformations
      across UNet training.
    ][t]{%
      \includegraphics[width=0.40\linewidth]{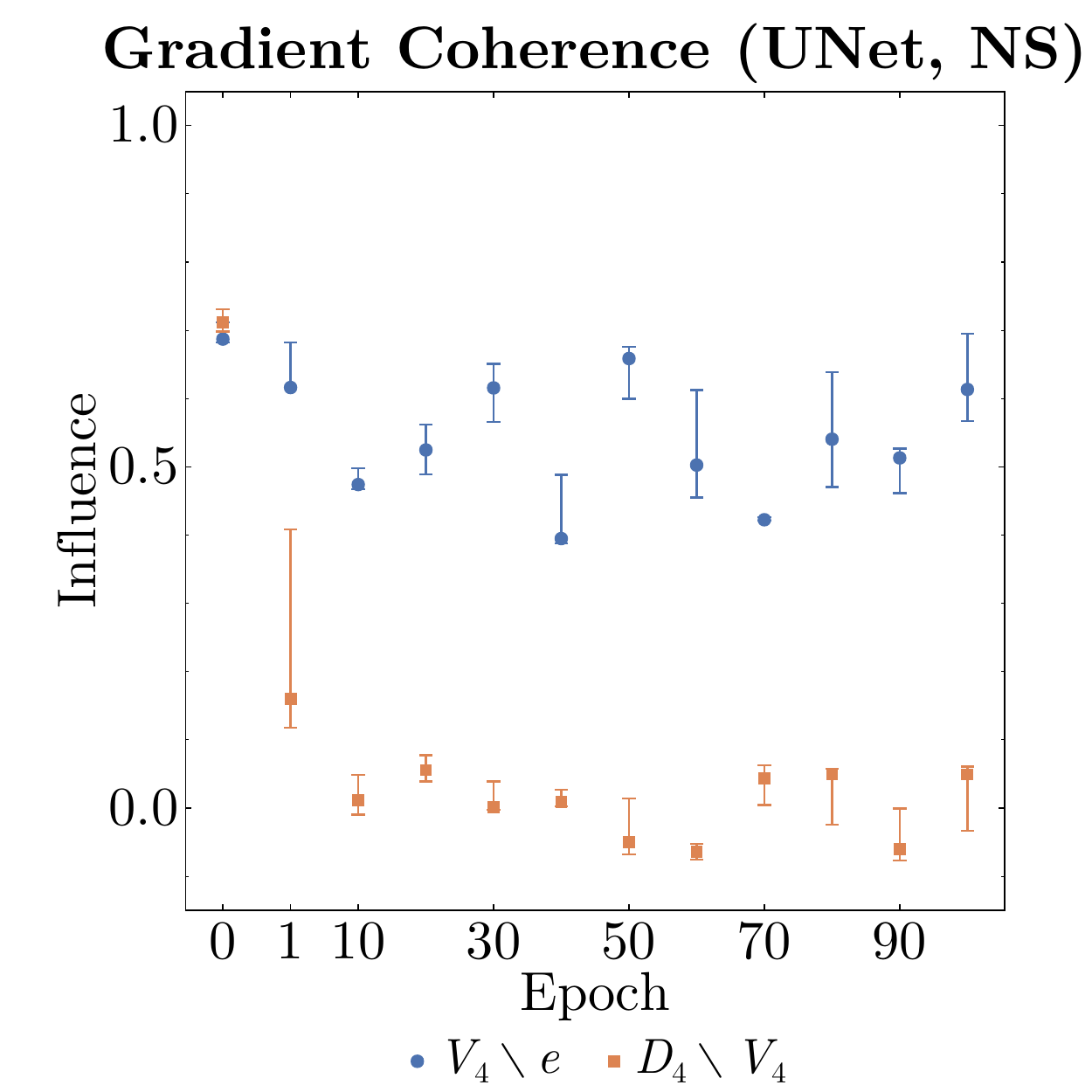}}
    \hfill
    \subfigure[
      Cross-influence into the learned and unlearned \(D_{4}\) transformations
      across ViT training.
    ][t]{%
      \includegraphics[width=0.40\linewidth]{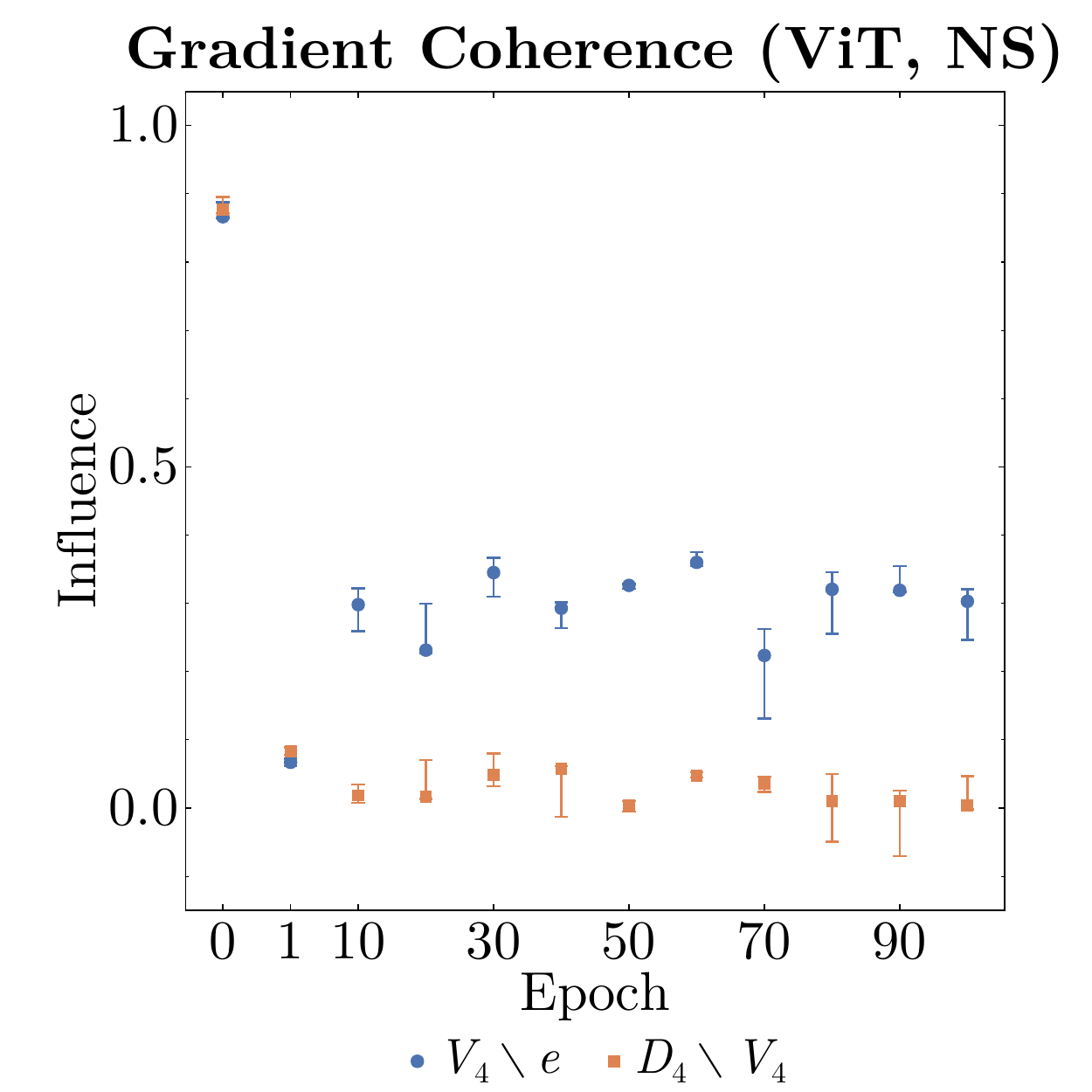}}
  }%
\end{figure}

\begin{figure}[htbp]
\floatconts
  {fig:D4_evolution_NS}
  {\caption{
    Evolution of cross-influence into each nonidentity \(D_{4}\) transformation
    for Navier-Stokes models. The elementwise profiles resolve the
    subgroup-averaged separation in \autoref{fig:Klein_NS}: \(r^{2}\), \(sr\),
    and \(sr^{3}\) in \(V_{4}\setminus\{e\}\) retain substantial
    cross-influence, whereas \(r\), \(r^{3}\), \(s\), and \(sr^{2}\) in
    \(D_{4}\setminus V_{4}\) rapidly become suppressed. This partition
    develops and persists for both architectures, showing that gradient
    transfer is organized according to the residual Klein four-subgroup
    rather than distributed across the full \(D_{4}\) orbit. Markers and
    rangebars indicate the median and interquartile range over seeds.
  }}
  {%
    \setlength{\jmlrminsubcaptionwidth}{0.40\linewidth}%
    \subfigure[
        Cross-influence into each nonidentity \(D_{4}\) transformation across UNet
  training.
    ][t]{%
      \includegraphics[width=0.40\linewidth]{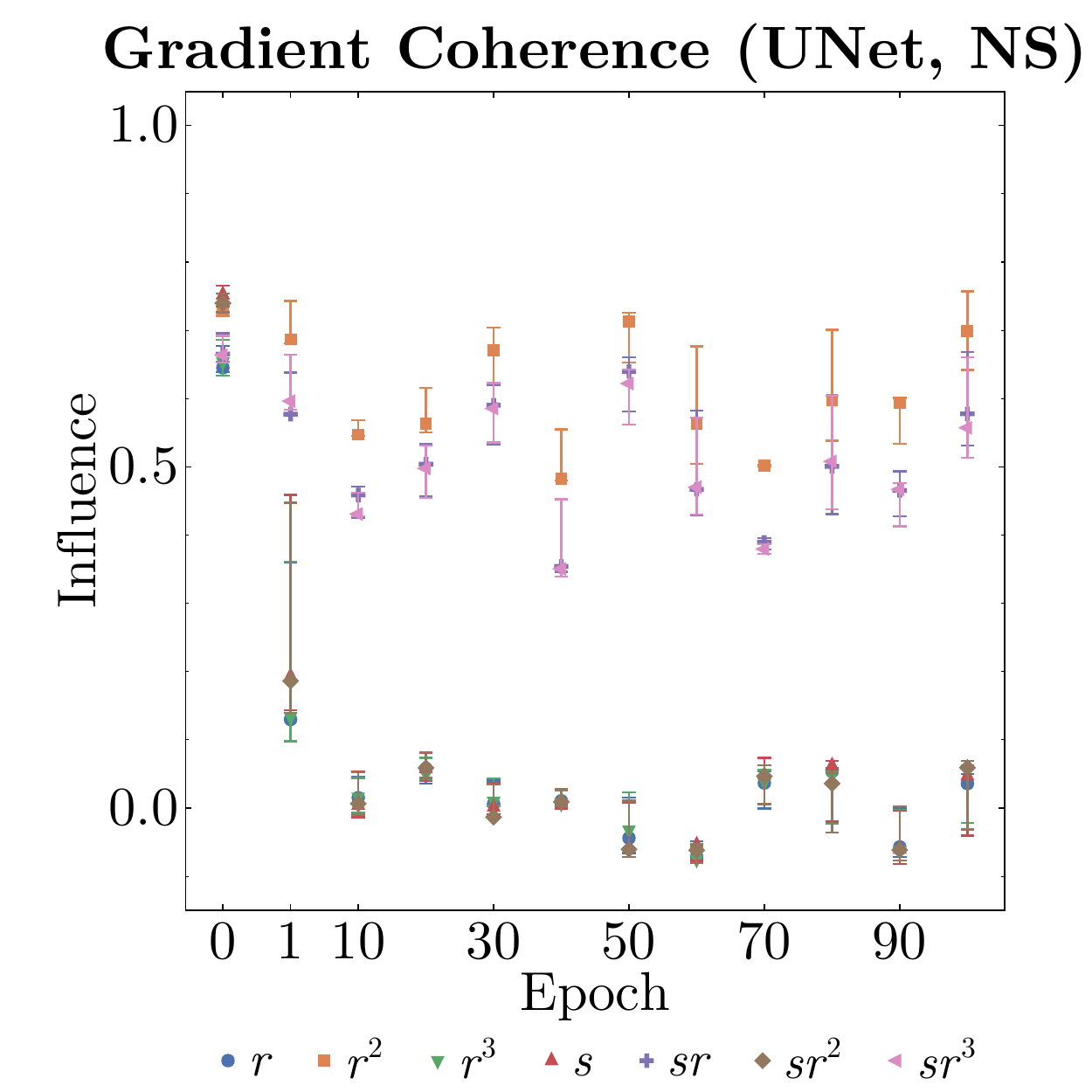}}
    \hfill
    \subfigure[
        Cross-influence into each nonidentity \(D_{4}\) transformation across ViT
  training.
    ][t]{%
      \includegraphics[width=0.40\linewidth]{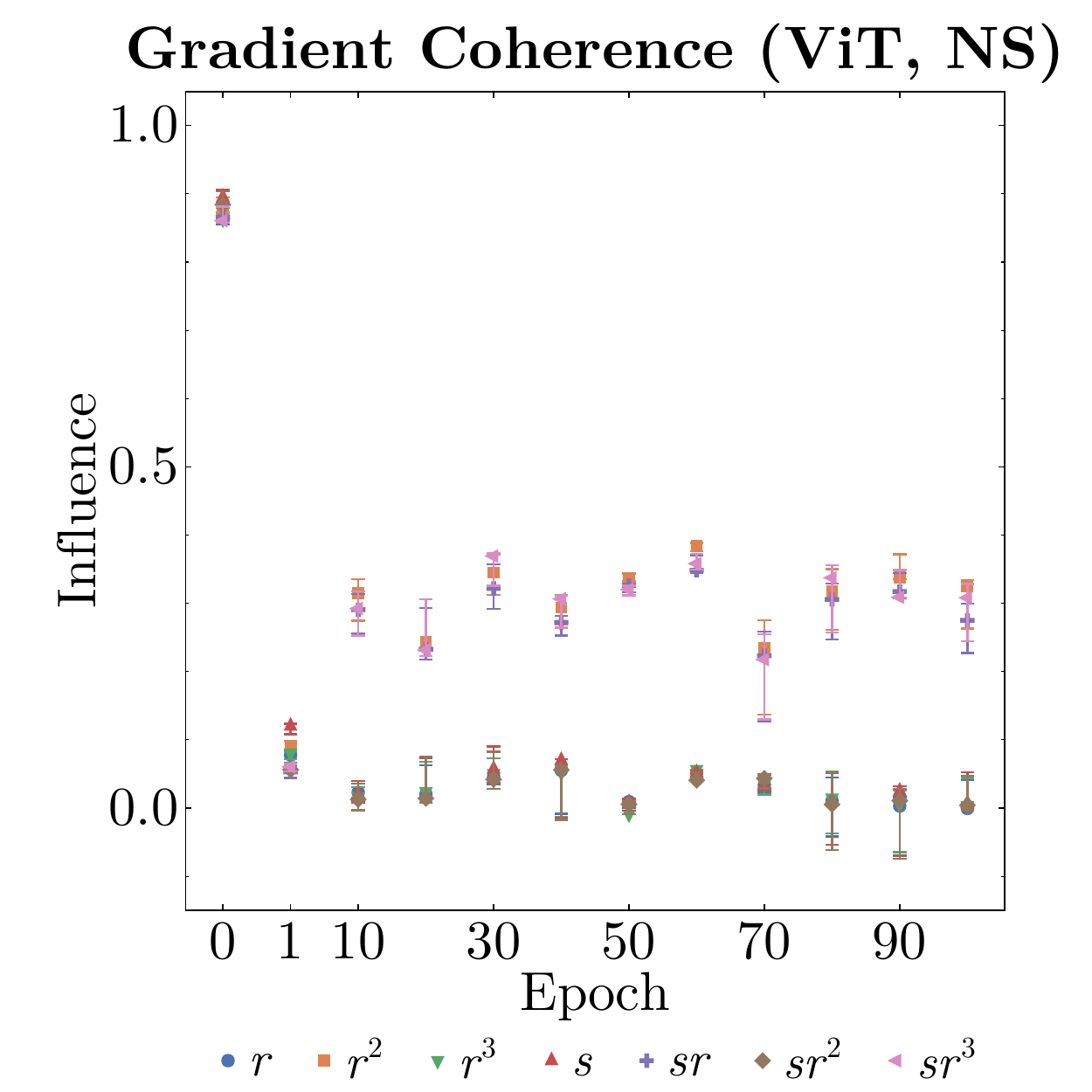}}
  }%
\end{figure}
\begin{figure}[htbp]
\floatconts
  {fig:D4_evolution_CE}
  {\caption{
              Evolution of cross-influence into each nonidentity \(D_{4}\) transformation
    for compressible Euler models. In contrast to the Navier-Stokes dynamics
    in \autoref{fig:D4_evolution_NS}, no persistent subgroup separation
    develops. Following the initial training transient, all seven nonidentity
    transformations retain positive cross-influence of comparable magnitude,
    so updates induced by in-distribution examples continue to reduce loss
    across the full \(D_{4}\) orbit. This elementwise coupling is consistent
    with the late-stage \(D_{4}\) equivariance shown in
    \autoref{fig:D4_CE}. Markers and rangebars indicate the median and
    interquartile range over seeds.
  }}
  {%
    \setlength{\jmlrminsubcaptionwidth}{0.40\linewidth}%
    \subfigure[
        Cross-influence into each nonidentity \(D_{4}\) transformation across UNet
  training.
    ][t]{%
      \includegraphics[width=0.40\linewidth]{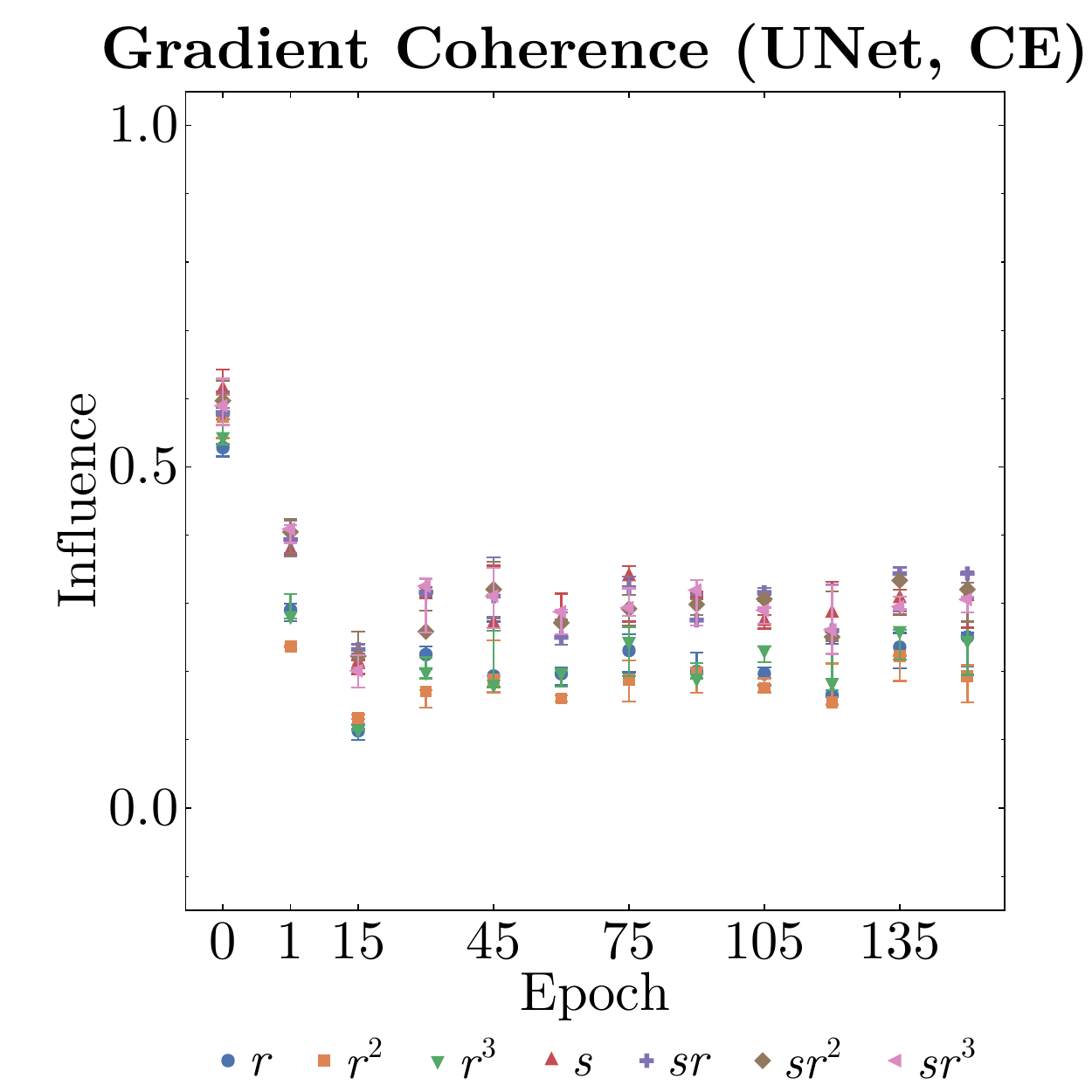}}
    \hfill
    \subfigure[
        Cross-influence into each nonidentity \(D_{4}\) transformation across ViT
  training.
    ][t]{%
      \includegraphics[width=0.40\linewidth]{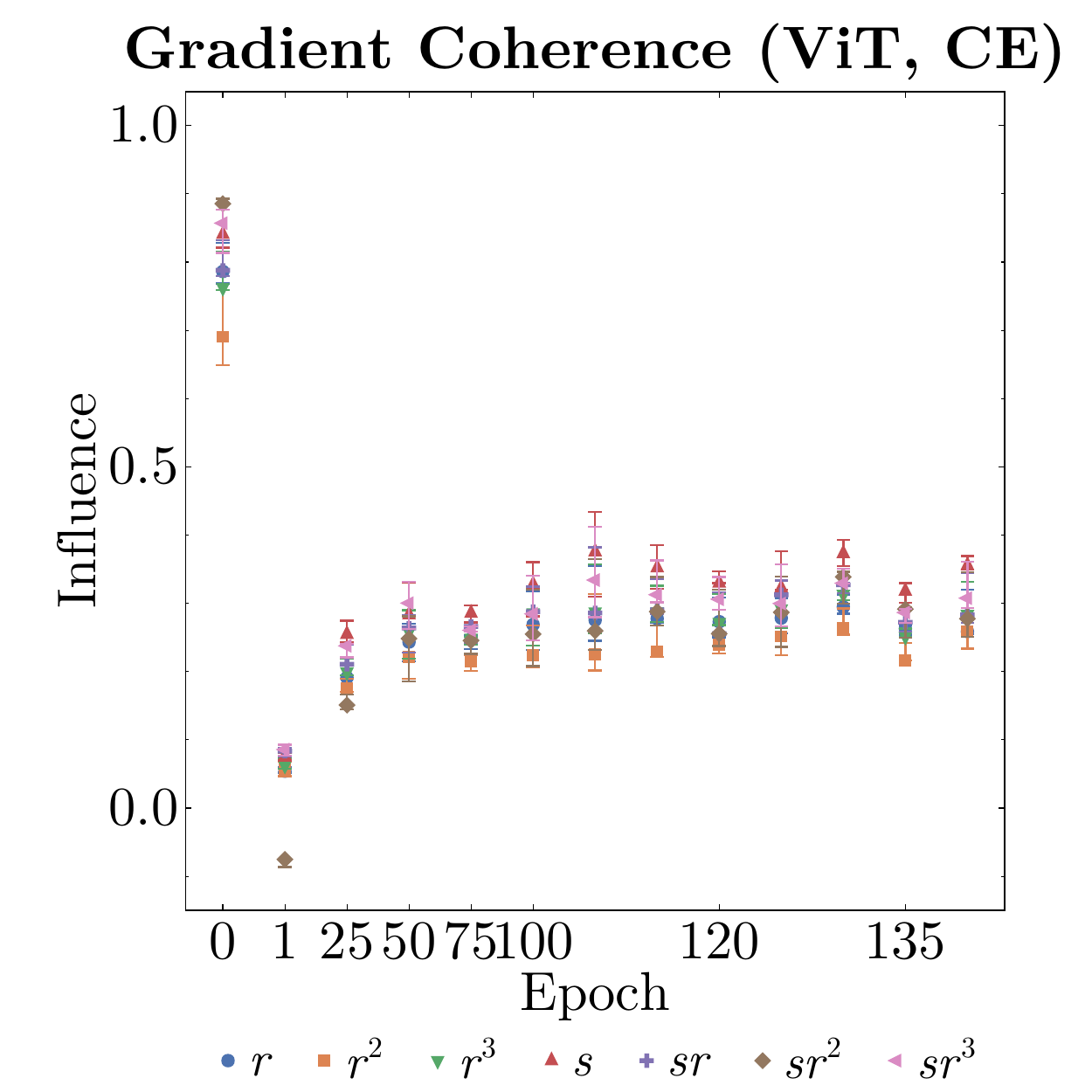}}
  }%
\end{figure}
\begin{figure}[htbp]
\floatconts
  {fig:box_CE_ViT}
  {\caption{
          Translation diagnostics for ViT on CE data over a sampled two-dimensional
translation grid. The forward-error and gradient-coherence landscapes exhibit
complementary phase structure: translations with elevated third-quartile
relative equivariance error lie in regions of reduced cross-influence. The
horizontal and vertical bands extend the axis-dependent resonances in
\autoref{fig:ZH_CE} and \autoref{fig:ZV_CE}. For the corresponding NS
landscape, see \autoref{fig:box_NS_ViT}; for the UNet response on CE data, see
\autoref{fig:box_CE_UNet}.
  }}
  {%
    \setlength{\jmlrminsubcaptionwidth}{0.40\linewidth}%
    \subfigure[
  Third-quartile relative equivariance error over joint horizontal and vertical
translations.
    ][t]{%
      \includegraphics[width=0.40\linewidth]{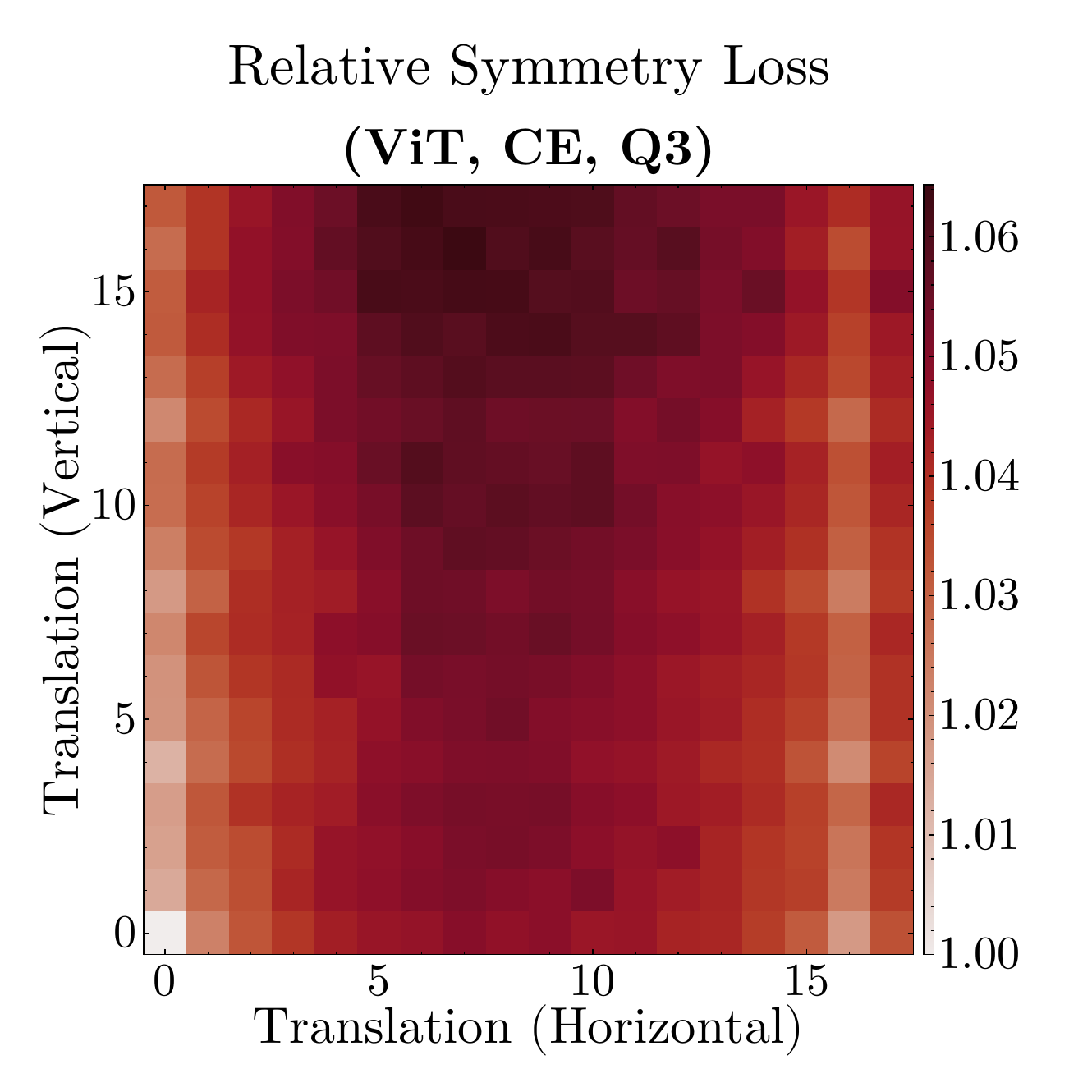}}
    \hfill
    \subfigure[
Cosine-normalized gradient alignment between an example and its jointly
translated state over the same grid.
    ][t]{%
      \includegraphics[width=0.40\linewidth]{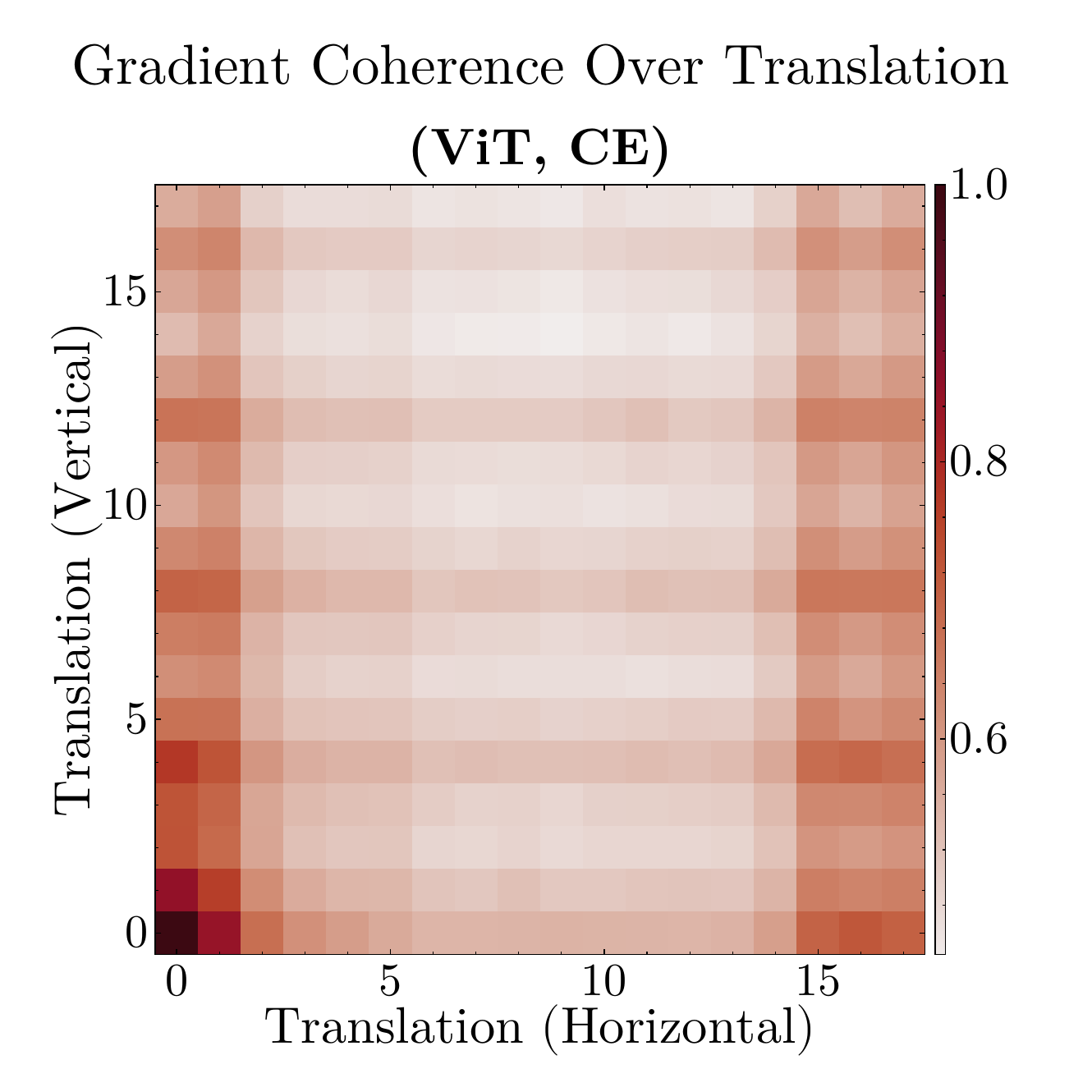}}
  }%
\end{figure}

\begin{figure}[htbp]
\floatconts
  {fig:box_CE_UNet}
  {\caption{
          Translation diagnostics for UNet on CE data over a sampled two-dimensional
translation grid. The error and coherence landscapes exhibit an approximately
eight-pixel lattice structure: relative-error minima along privileged
translation phases coincide with maxima in gradient coherence, whereas the
intervening regions exhibit larger error and weaker coupling. This
two-dimensional periodicity is consistent with the multiscale UNet response in
\autoref{fig:ZH_CE} and \autoref{fig:ZV_CE}. For the corresponding ViT
landscape, see \autoref{fig:box_CE_ViT}; for the UNet response on NS data, see
\autoref{fig:box_NS_UNet}.
  }}
  {%
    \setlength{\jmlrminsubcaptionwidth}{0.40\linewidth}%
    \subfigure[
  Third-quartile relative equivariance error over joint horizontal and vertical
translations.
    ][t]{%
      \includegraphics[width=0.40\linewidth]{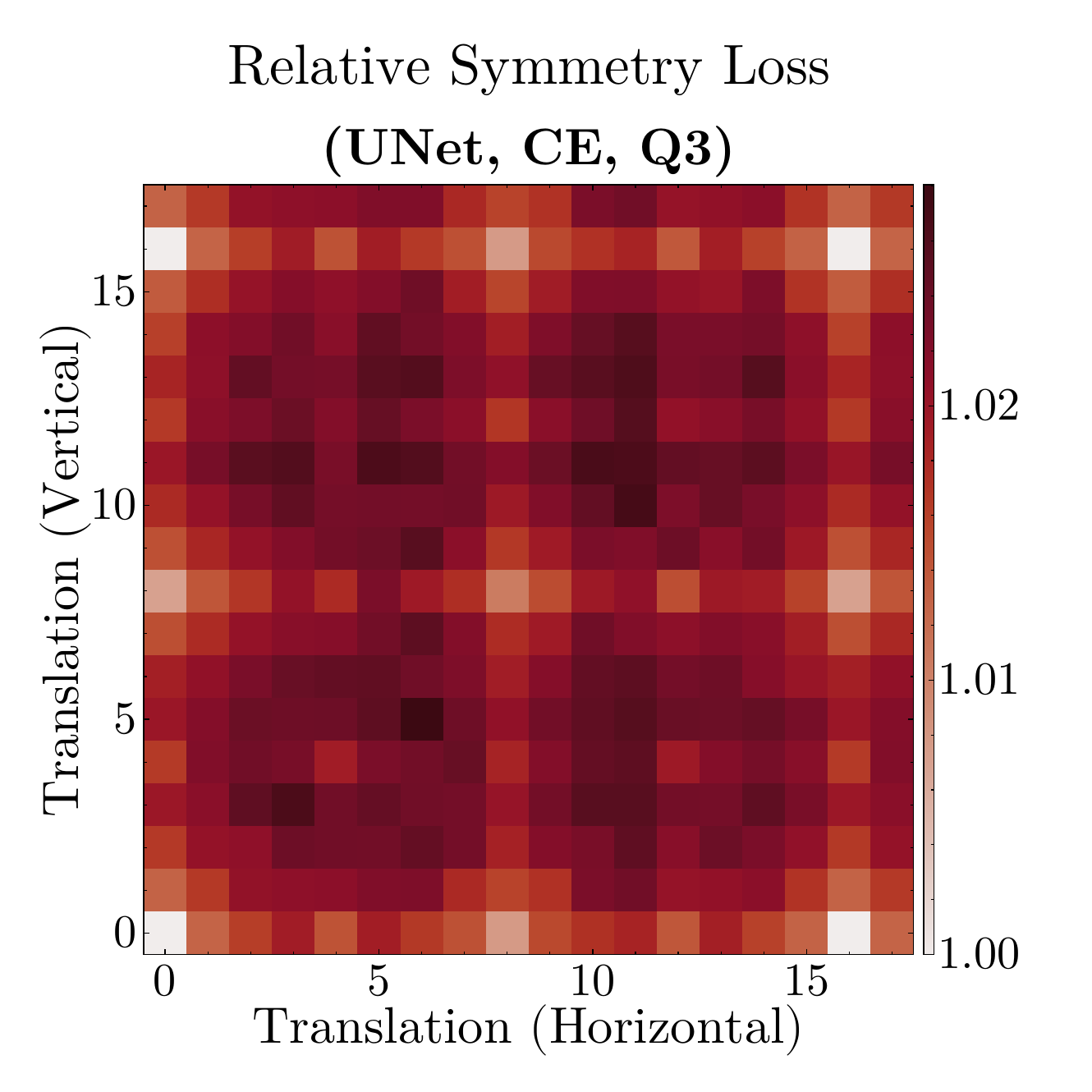}}
    \hfill
    \subfigure[
Cosine-normalized gradient alignment between an example and its jointly
translated state over the same grid.
    ][t]{%
      \includegraphics[width=0.40\linewidth]{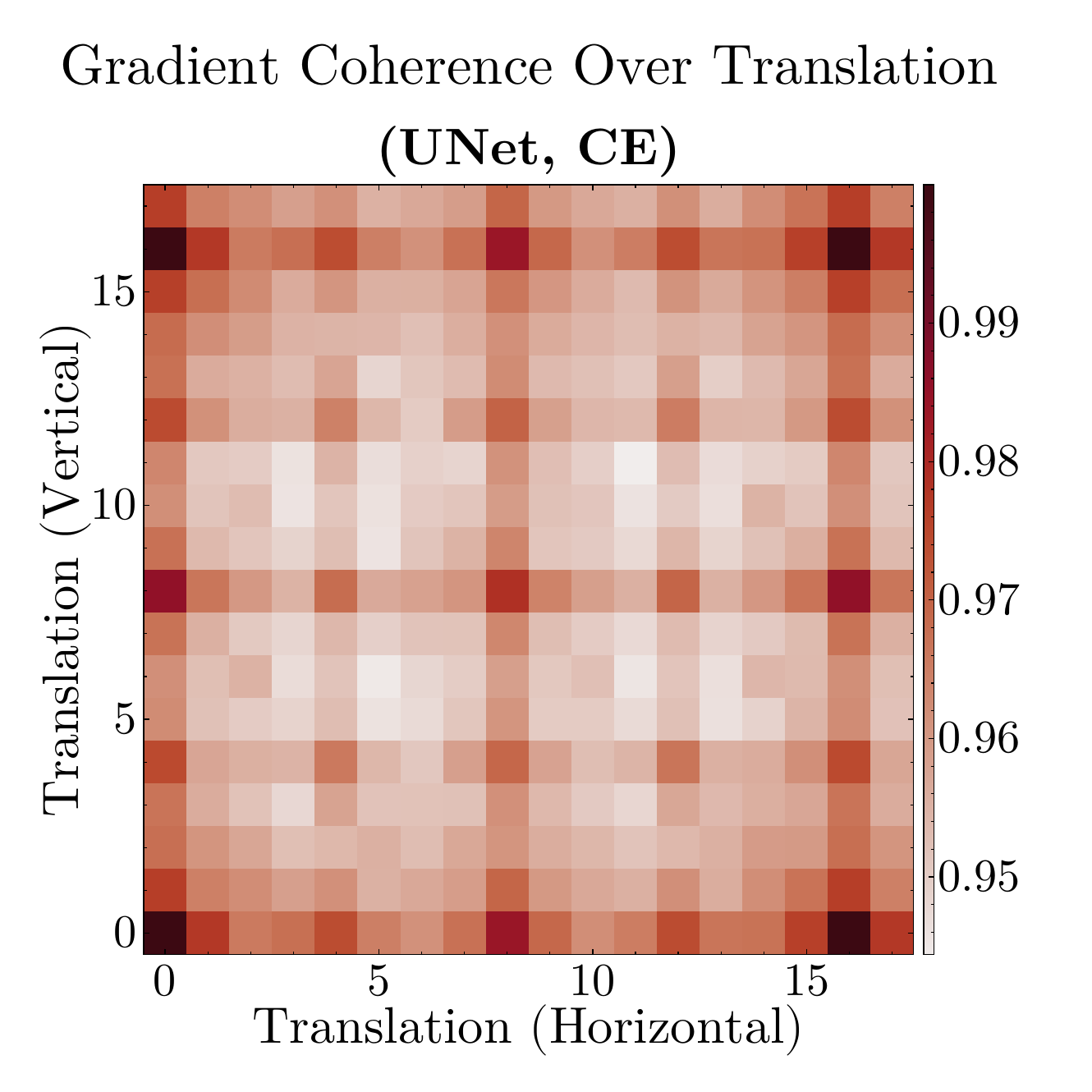}}
  }%
\end{figure}

\begin{figure}[htbp]
\floatconts
  {fig:box_NS_ViT}
  {\caption{
Translation diagnostics for ViT on Navier--Stokes (NS) data over a sampled
two-dimensional translation grid. This case exhibits the weakest structural
correspondence between relative equivariance error and gradient coherence among
the translation measurements. Third-quartile relative equivariance error has a
banded dependence on translation phase, whereas gradient coherence decays
approximately isotropically with distance from the identity translation and
does not reproduce the band structure of the error landscape. The corresponding
one-dimensional measurements appear in \autoref{fig:ZH_NS} and
\autoref{fig:ZV_NS}. For the corresponding UNet landscape, see
\autoref{fig:box_NS_UNet}; for the ViT response on CE data, see
\autoref{fig:box_CE_ViT}.
  }}
  {%
    \setlength{\jmlrminsubcaptionwidth}{0.40\linewidth}%
    \subfigure[
  Third-quartile relative equivariance error over joint horizontal and vertical
translations.
    ][t]{%
      \includegraphics[width=0.40\linewidth]{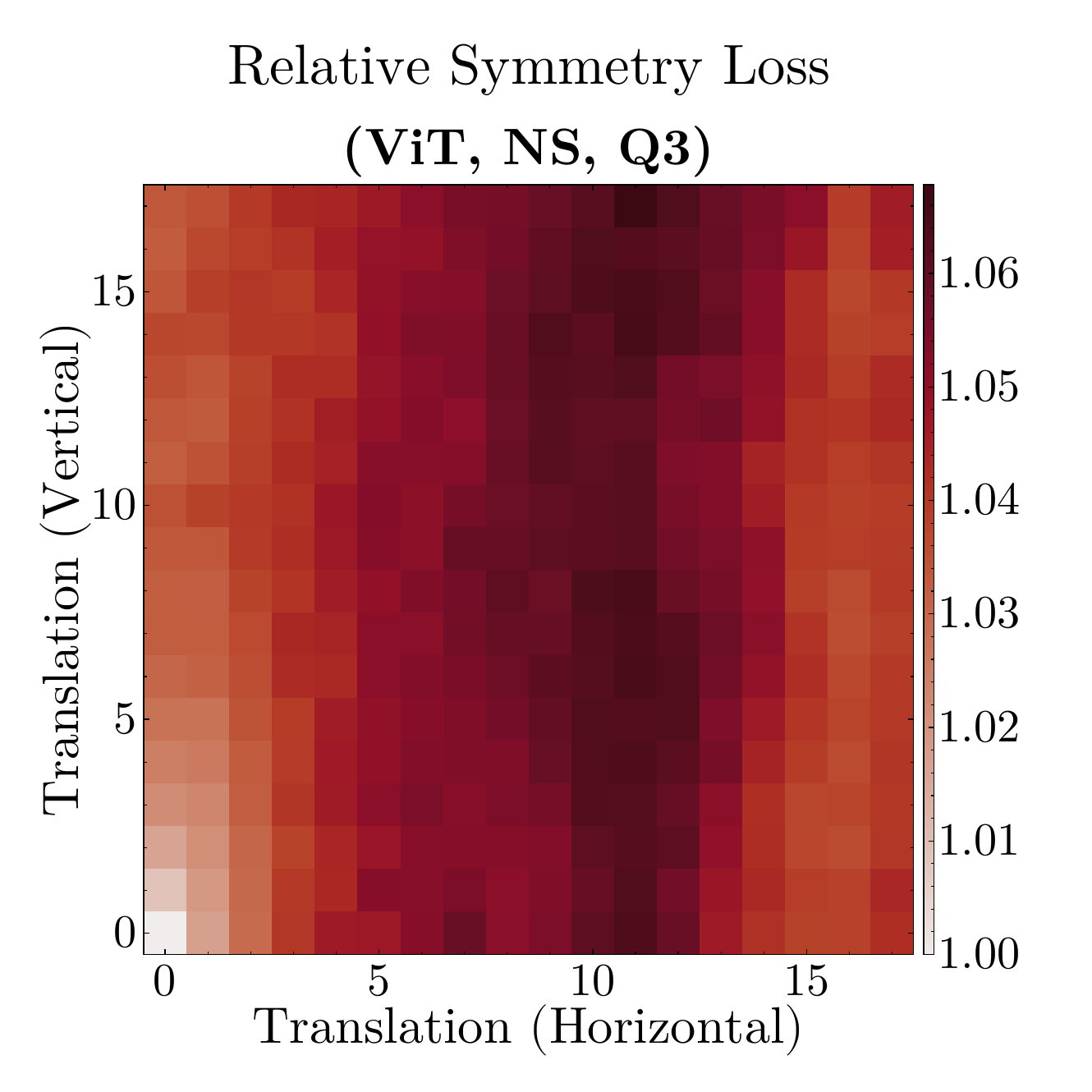}}
    \hfill
    \subfigure[
Cosine-normalized gradient alignment between an example and its jointly
translated state over the same grid.
    ][t]{%
      \includegraphics[width=0.40\linewidth]{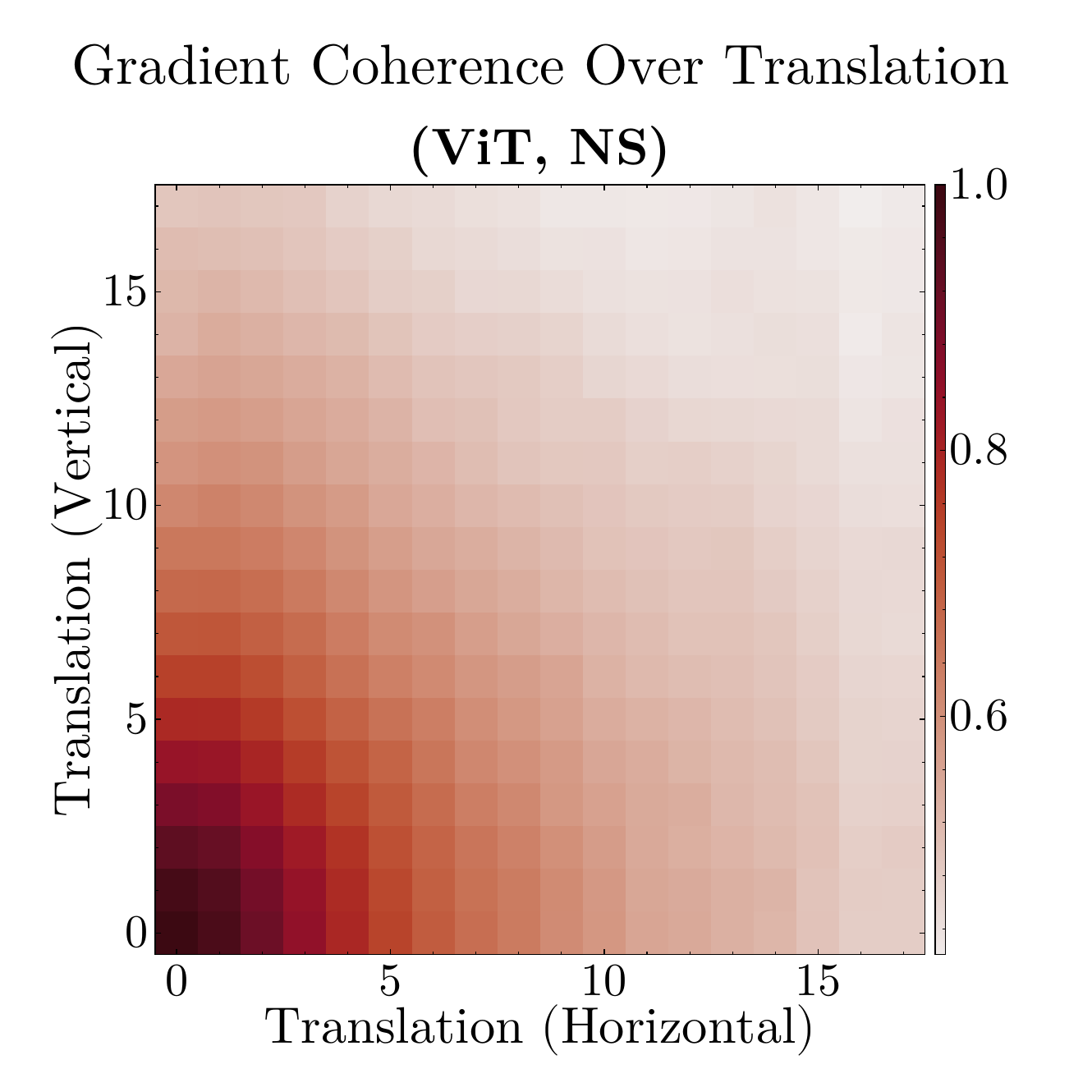}}
  }%
\end{figure}

\begin{figure}[htbp]
\floatconts
  {fig:box_NS_UNet}
  {\caption{
          Translation diagnostics for UNet on Navier--Stokes (NS) data over a sampled
two-dimensional translation grid. Gradient coherence remains close to the unit
self-response throughout the grid, with a smooth central suppression that
corresponds to elevated third-quartile relative equivariance error. Relative to
the ViT landscape in \autoref{fig:box_NS_ViT}, the UNet distributes gradient
coherence more uniformly across joint translations. For the corresponding CE
landscape, see \autoref{fig:box_CE_UNet}.
  }}
  {%
    \setlength{\jmlrminsubcaptionwidth}{0.40\linewidth}%
    \subfigure[
  Third-quartile relative equivariance error over joint horizontal and vertical
translations.
    ][t]{%
      \includegraphics[width=0.40\linewidth]{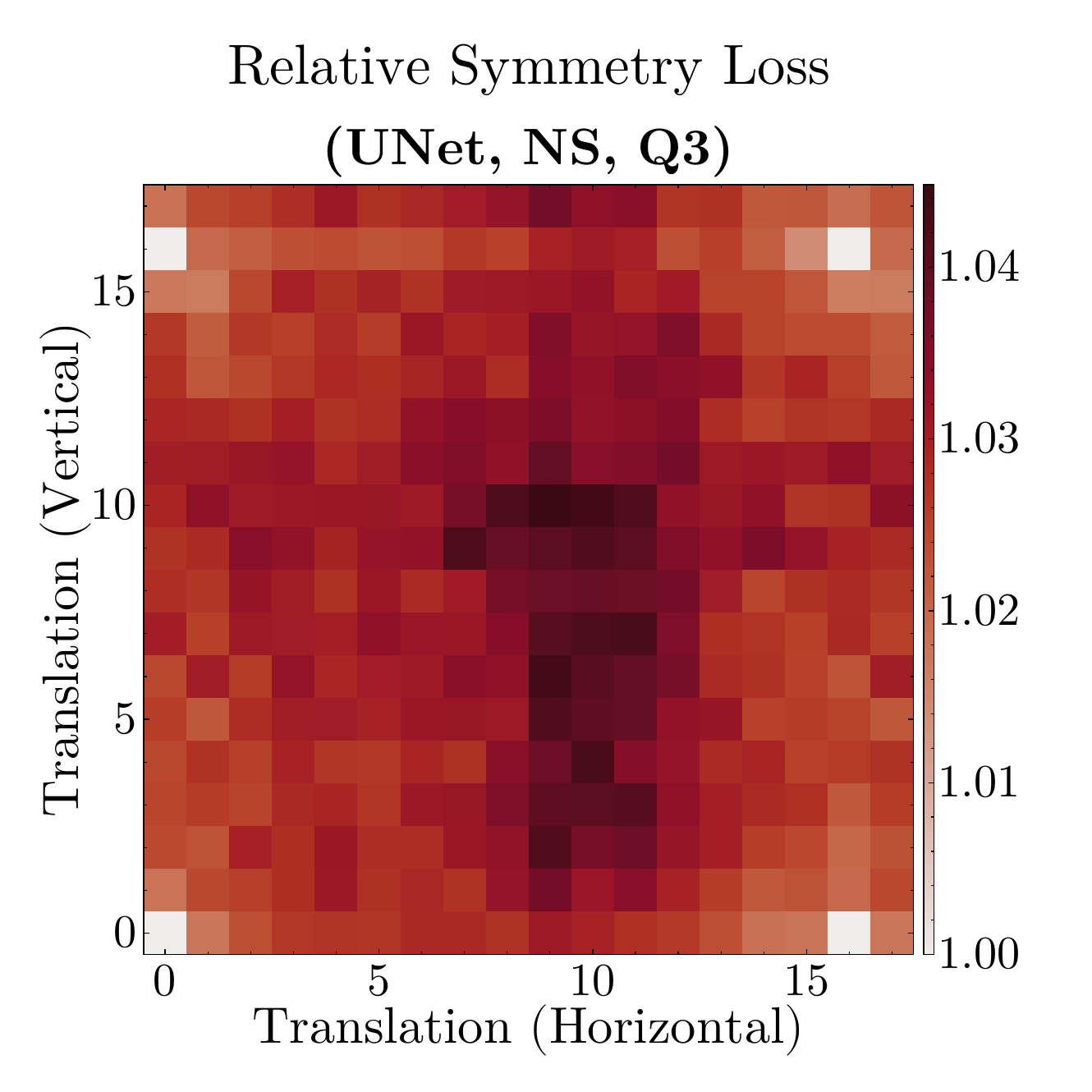}}
    \hfill
    \subfigure[
Cosine-normalized gradient alignment between an example and its jointly
translated state over the same grid.
    ][t]{%
      \includegraphics[width=0.40\linewidth]{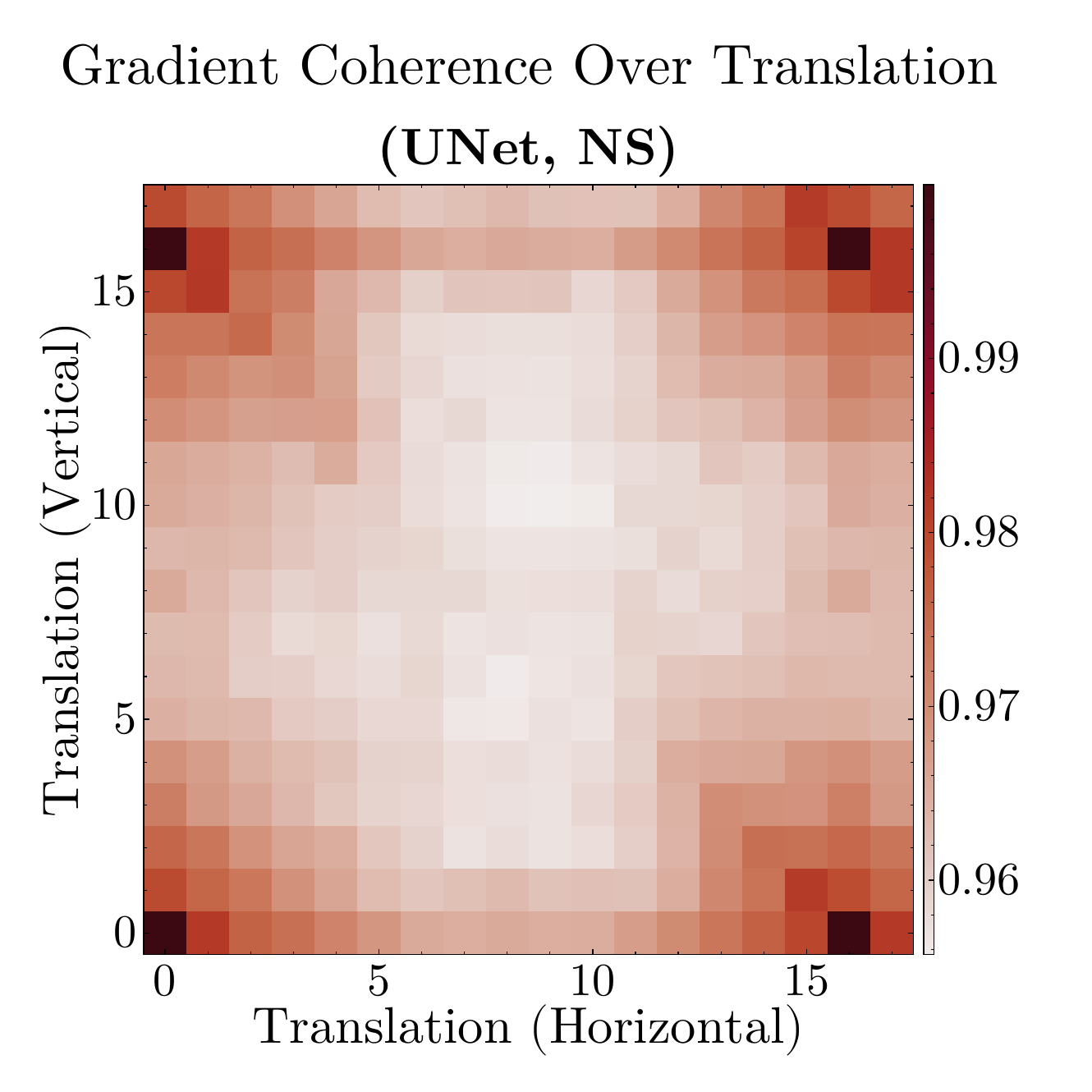}}
  }%
\end{figure}

\begin{figure}[htbp]
\floatconts
  {fig:opt_curves}
  {\caption{Optimization trajectories for the CE and NS tasks.
          Test scaled mean squared error (SMSE) is shown as a function of training epoch for UNets and ViTs.
          Markers denote median performance across seeds, and rangebars indicate variability across seeds at each epoch.
          The curves establish the optimization regimes corresponding to the late-stage
symmetry diagnostics and the checkpointed influence measurements.
  }}
  {%
    \setlength{\jmlrminsubcaptionwidth}{0.40\linewidth}%
    \subfigure[
    CE optimization trajectories.
    ][t]{%
      \includegraphics[width=0.40\linewidth]{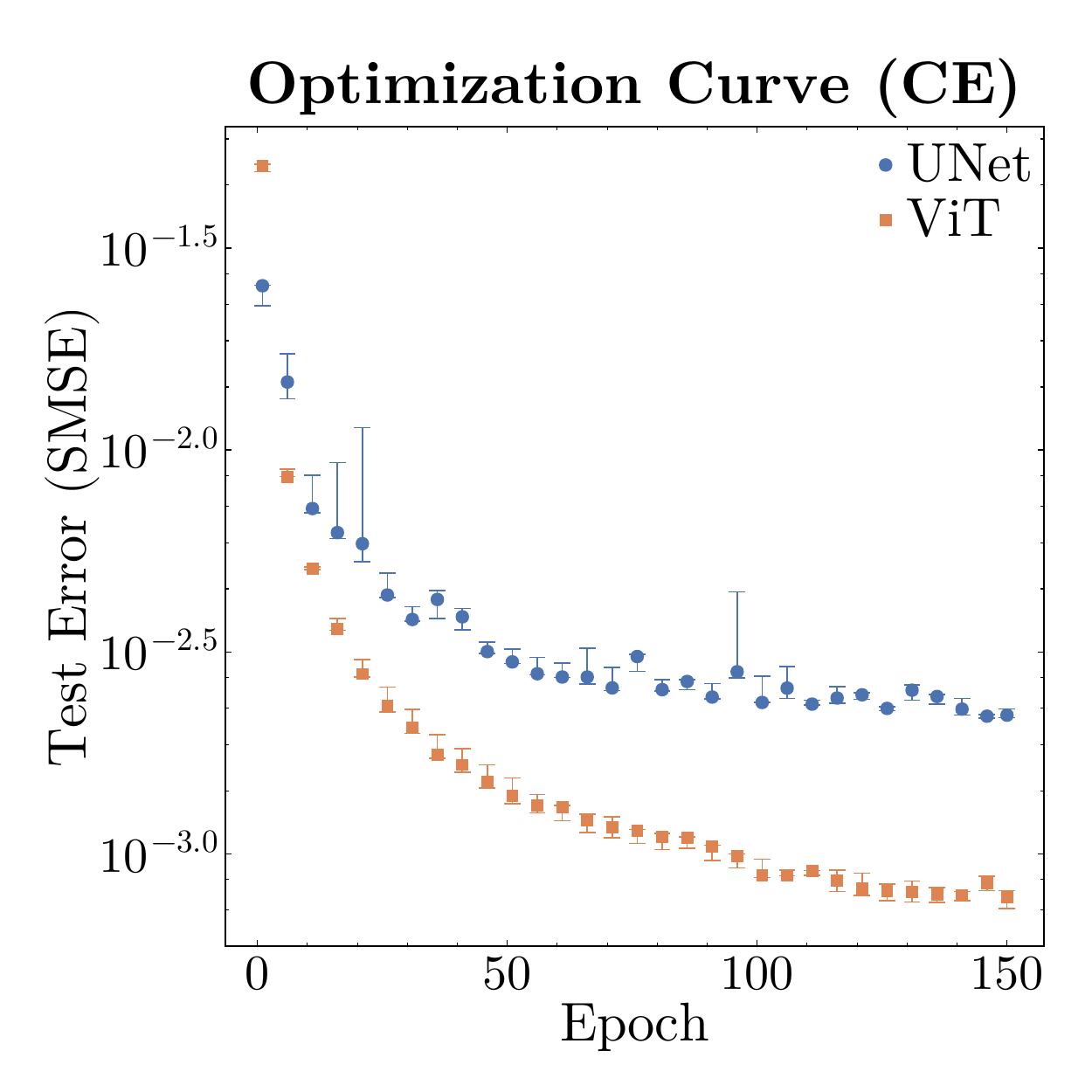}}%
    \hfill
    \subfigure[
    NS optimization trajectories.
    ][t]{%
      \includegraphics[width=0.40\linewidth]{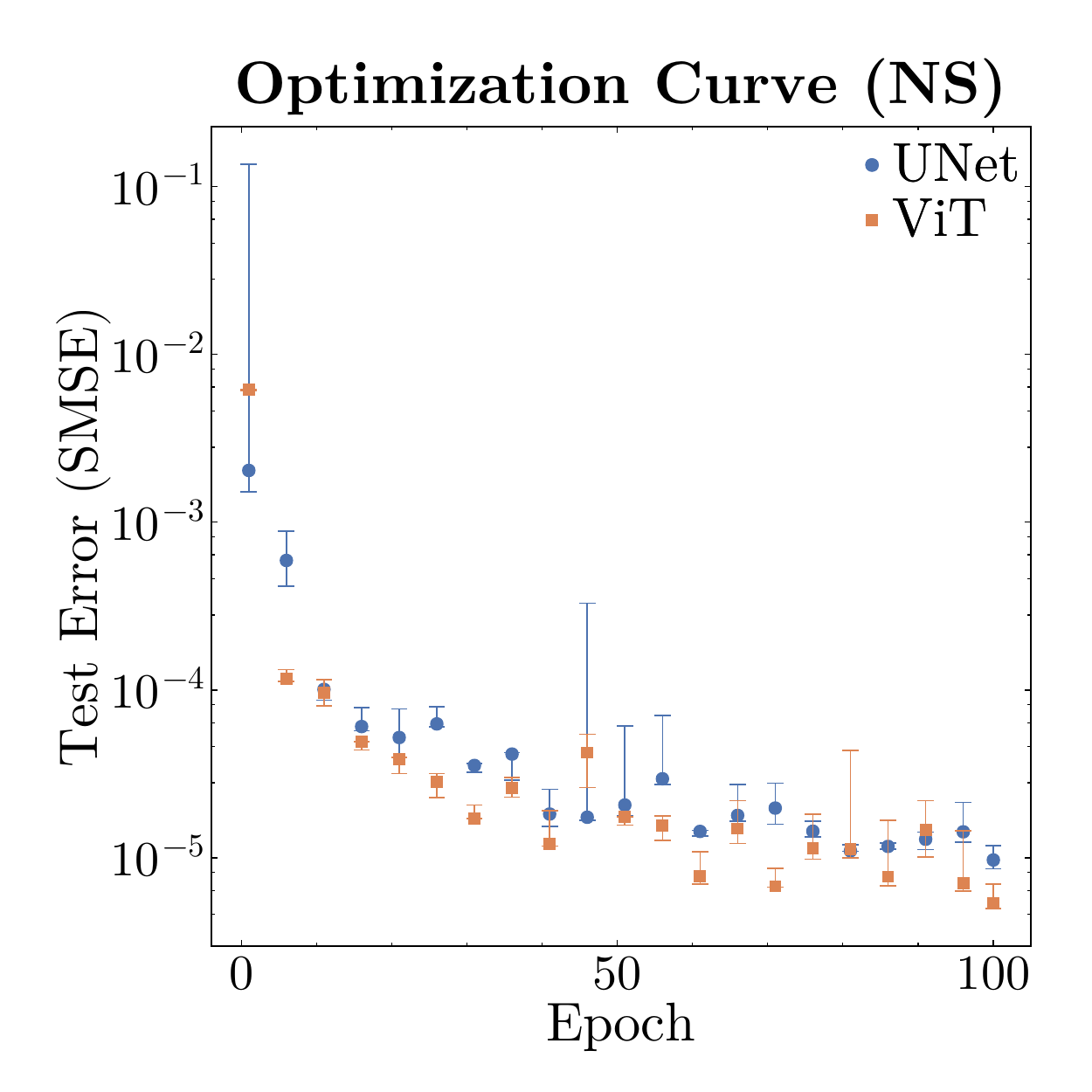}}
  }%
\end{figure}

\end{document}